\documentclass[twocolumn]{svjour3} 

\usepackage[linesnumbered,ruled]{algorithm2e}
\usepackage{subfigure}
\usepackage{amsmath}
\usepackage{graphicx}
\usepackage{float}
\usepackage{pifont}
\usepackage{amssymb}
\usepackage{xfrac}
\usepackage{lipsum}
\usepackage{amsfonts}
\usepackage{booktabs}
\usepackage{enumitem}

\usepackage[export]{adjustbox}

\usepackage[authoryear]{natbib}
\newtheorem{mydef}{Definition}
\newcommand{\argmax}{\arg\!\max}

\begin{document}
	
\title{Geometry of Interest (GOI): Spatio-Temporal Destination Extraction and Partitioning in GPS Trajectory Data}
\author{Seyed Morteza Mousavi \and Aaron Harwood \and Shanika Karunasekera \and Mojtaba Maghrebi}
\institute{\at
    NICTA VRL, Department of Computing and Information Systems, The University of Melbourne, Australia \\
    Tel.: +61-405967448\\
    \email{mousavi@student.unimelb.edu.au}
    \and
    \at
    Department of Computing and Information Systems, The University of Melbourne, Australia \\
    Tel.: +61-383441351\\
    \email{aharwood@unimelb.edu.au}
    \and
    \at
    Department of Computing and Information Systems, The University of Melbourne, Australia \\
    Tel.: +61-383441351\\
    \email{karus@unimelb.edu.au}
    \and
    \at
    School of Civil and Environmental Engineering, UNSW, Australia \\
    Tel.: +61-383441351\\
    \email{maghrebi@unsw.edu.au}
}

\date{Received: date / Accepted: date}

\maketitle

\begin{abstract}
	Nowadays large amounts of GPS trajectory data is being continuously collected by GPS-enabled devices such as vehicles navigation systems and mobile phones. GPS trajectory data is useful for applications such as traffic management, location forecasting, and itinerary planning. Such applications often need to extract the time-stamped Sequence of Visited Locations (SVLs) of the mobile objects. The nearest neighbor query (NNQ) is the most applied method for labeling the visited locations based on the IDs of the POIs in the process of SVL generation. NNQ in some scenarios is not accurate enough. To improve the quality of the extracted SVLs, instead of using NNQ, we label the visited locations as the IDs of the POIs which geometrically intersect with the GPS observations. Intersection operator requires the accurate geometry of the points of interest which we refer to them as the Geometries of Interest (GOIs). In some application domains (e.g. movement trajectories of animals), adequate information about the POIs and their GOIs may not be available a priori, or they may not be publicly accessible and, therefore, they need to be derived from GPS trajectory data. In this paper we propose a novel method for estimating the POIs and their GOIs, which consists of three phases: (i) extracting the geometries of the stay regions; (ii) constructing the geometry of destination regions based on the extracted stay regions; and (iii) constructing the GOIs based on the geometries of the destination regions.  Using the geometric similarity to known GOIs as the major evaluation criterion, the experiments we performed using long-term GPS trajectory data show that our method outperforms the existing approaches.
\end{abstract}

\section{Introduction}
\label{introduction}

\begin{sloppypar}
	
	In recent years, GPS trajectory data has become abundant due to the many GPS enabled devices used on a daily basis. Mining these GPS trajectories for gathering useful information for applications has received a growing amount of attention in the recent literature. In this field, researchers have tried to derive knowledge for solving practical problems (e.g. traffic and transportation management systems~\citep{trafficpredition:2010}, animal migration and movement monitoring~\citep{AnimalMonitoring:2009}, location prediction~\citep{Falvi:WhenandWhereNext:2012}, transportation mode estimation~\citep{Zheng:TransportationMode:2010}, and location-based social networks~\citep{MobileIntelligence:2011}). 
	
	The applications dealing with data analysis on trajectory data often need to have access to information about the significant places which a mobile object frequently travels and stay. These significant places are referred to as the points of interest (POIs). The locations of the POIs are often used in projecting the trajectory of a mobile object into a meaningful time-stamped Sequence of Visited Locations (SVL). The constructed sequences are used in various machine learning applications dealing with trajectory data~\citep{yan:semantic2011}. Therefore, the quality and accuracy of the sequence have very significant impact on the performance of the machine learning applications.
	
\begin{figure}
		\centering
		\subfigure[t][Nearest Neighbor Based]{\label{fig:MotivationDestinations}\includegraphics[width=0.45\linewidth]{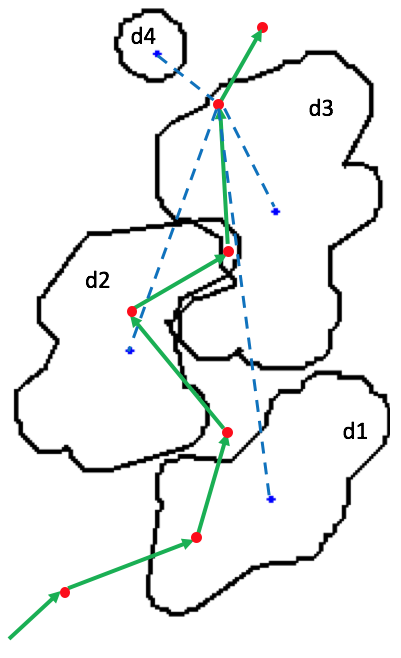}}
		~\hspace{1em}
		\subfigure[t][GOI Based]{\label{fig:MotivationGOIs}\includegraphics[width=0.45\linewidth]{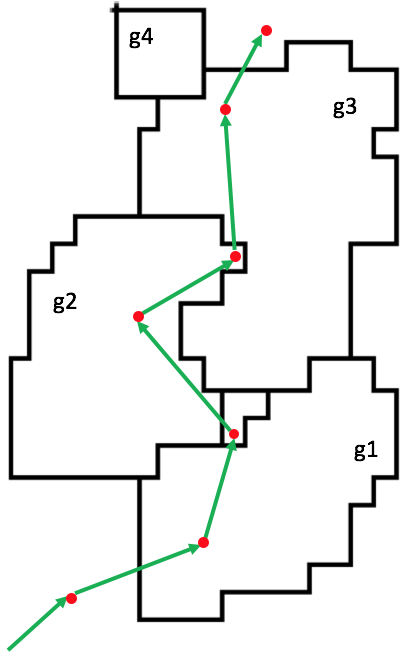}}
		%~\hspace{1em}
		%\subfigure[t][Final-Grid Based Labeling]{\label{fig:MotivationFinal}\includegraphics[width=0.30\linewidth]{Motivation_Final.png}}
		\centering
		\caption{Labling Approaches in the Process of Generating the Sequence of Visited Locations (SVL)}
		\label{fig:Motivation}
	\end{figure}

	In the process of constructing the SVL of a trajectory, the applications often use Nearest Neighbor Queries (NNQ) to label each GPS observation with the ID of a POI. Fig.~\ref{fig:MotivationDestinations} presents an overview of the SVL construction process. Given a GPS trajectory (depicted with green arrows), and a set of destinations showed as polygons, the NNQ based labeling method labels each GPS observation (depicted in red points) with the ID of the nearest POI (centroid of the destination polygons) in chronological order. Although this process is quite simple and efficient, it has a few significant limitations which have a dramatic impact on the quality of the generated time-stamped SVL. 
	
	As an example, consider the GPS point covered by destination $d3$. The labeling process labels the destination with the ID of the nearest destination $d4$, while in the real world, the GPS coordinate intersect with the geometry of $d3$. Therefore, the resulting SVL for the depicted trajectory using NNQ based labeling method is $d1 \rightarrow d1 \rightarrow d1 \rightarrow d2 \rightarrow d3 \rightarrow d4$. This scenario frequently happens especially in environments with a high number of POIs located near each other. The problem has a dramatic impact of the quality of the constructed SVLs.
	
	One solution to the problem is to label each of the GPS points with the ID of the POIs which \emph{intersect} with them instead of performing NNQ. Fig.~\ref{fig:MotivationGOIs}, shows the estimated geometry of each of the POIs which we refer to them as the Geometries of Interest (GOIs). Moreover, the GOIs must not be overlapping. Otherwise, the intersection operator would not be able to label a GPS point intersecting with more than one GOI. This solution requires having access to the real world GOIs stored in a spatial database. 
	
	The information about the GOIs might be publicly available in the spatial databases (e.g. geometries of the famous places in a city). However, in applications such as those processing the motion patterns of animals or the movement patterns of the troops in a battlefield, GOIs are not available and are required to be extracted from the trajectory data.
	
	In this paper, we address the problem of extracting the GOIs of a mobile object, without using any information other than the GPS trajectory of the mobile object. We propose a method to partition the trajectory area, which is defined by the minimum bounding rectangle (MBR) of the trajectory, into a grid containing the GOIs of the moving object. Using the extracted GOIs and the partitioned trajectory area, we can extract the trajectory SVL by only using intersection geometric operator. The quality and accuracy of the SVL highly depend on the accuracy of the estimated GOIs. 
	
	Aiming for that, we extend the spatio-temporal partitioning techniques proposed in~\citep{Zheng:LifePatterns:2009,ProjectLachesis:2004}. The partitioning methods have three phases. Firstly, they extract the stay regions within which a moving object has stayed for a time duration greater or equal than a predefined minimum time threshold and within a predefined Euclidean vicinity distance. Secondly, they cluster the resulting stay points (the centroids of the stay regions) to extract the destinations of the moving objects. Thirdly, they implicitly partition the trajectory area based on the coordinates of the centroids of the extracted destinations by using NNQ in the process of labeling the GPS points with the identifications of the POIs. 
	
	\begin{figure*}
		\centering
		\subfigure[t][GPS Trajectory]{\label{fig:OveralTrajectory}\includegraphics[width=0.22\linewidth,frame]{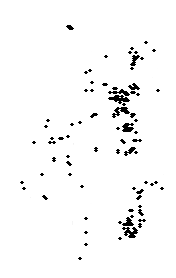}}
		~\hspace{1em}
		\subfigure[t][Extracted Stay Regions (First Phase)]{\label{fig:OveralStayRegions}\includegraphics[width=0.22\linewidth]{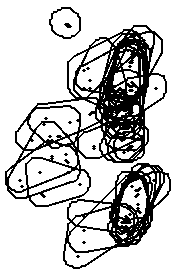}}
		~\hspace{1em}
		\subfigure[t][Extracted Destination Regions (Second Phase)]{\label{fig:OveralDestinationRegions}\includegraphics[width=0.22\linewidth]{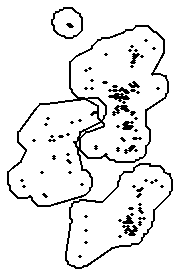}}
		~\hspace{1em}
		\subfigure[t][Final Partitioned Area (Third Phase)]{\label{fig:OveralFinalGrid}\includegraphics[width=0.22\linewidth]{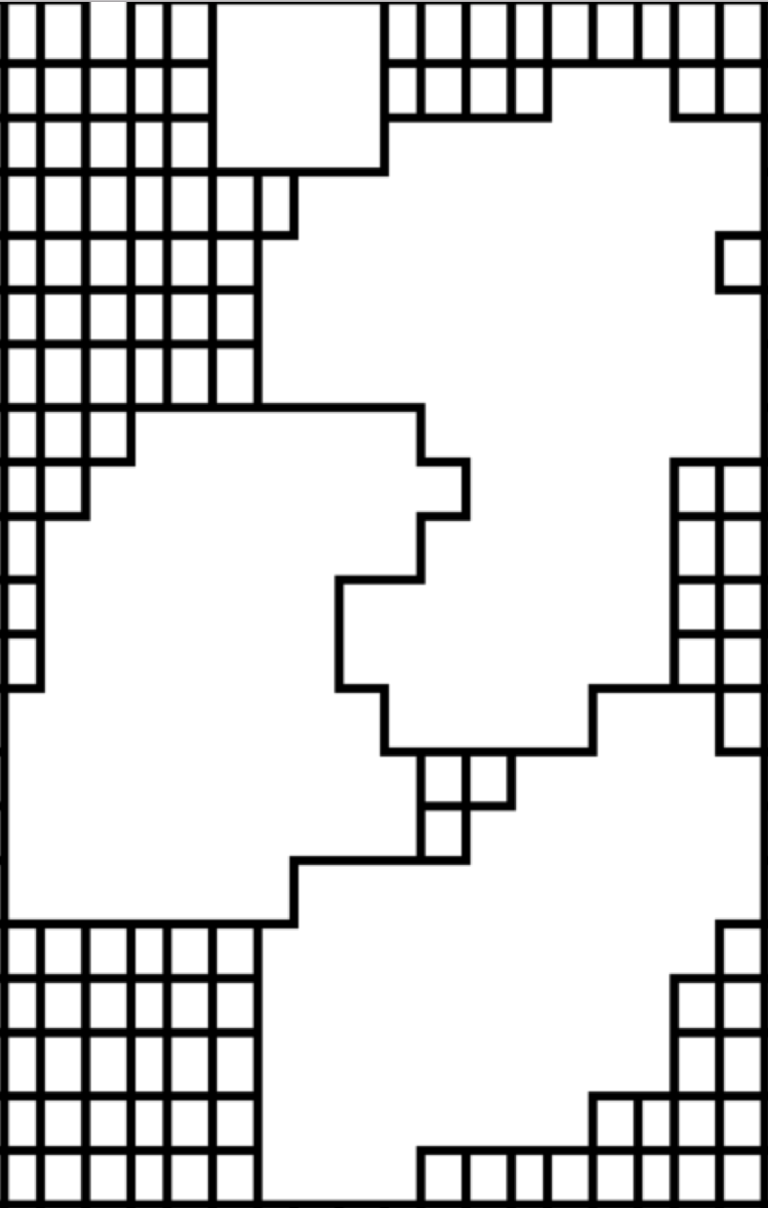}}
		\centering
		\caption{The Results of the Spatio-Temporal Partitioning Phases.}
		\label{fig:ResultsoftheSpatio-TemporalPartitioningPhases}
	\end{figure*}
	
	Our proposed method improves the baselines, in each of the three phases. Given a GPS trajectory (Fig.~\ref{fig:OveralTrajectory}), in the stay extraction phase, we propose a novel clustering method for constructing the stay regions (Fig.~\ref{fig:OveralStayRegions}). In the destination construction phase, we propose a geometry based hierarchical agglomerative clustering method for clustering (merging) the stay regions based on a geometric similarity measure and construct the geometries of the destinations (Fig.~\ref{fig:OveralDestinationRegions}). In the third phase, we extract the GOIs based on the geometries of the destinations and include them in the final grid which is composed of the GOIs and the cells with fixed sizes (Fig.~\ref{fig:OveralFinalGrid}).
	
	The performance of our approach is evaluated based on comparing the similarity of the derived GOIs from our approach to know geometries of the POIs. Our experimental results performed on a long-term GPS trajectories show that, in the stay extraction phase, our method outperforms the existing methods by making the higher number of valid stay regions with geometries more related to the real world POIs. In the destination extraction phase, the performance and the accuracy of our method are considerably higher than the baseline methods, considering the geometric similarity between the geometries of the extracted destination to the real world POIs. Moreover, our method is able to partition the trajectory area based on the extracted destinations resulting in a grid which guarantees the characteristics of a validly partitioned area. Using the resulting grid, we can easily generate the SVL of the mobile object by using intersection geometric operator instead of using the nearest neighbor queries or Voronoi diagrams~\citep{Voronoi:1991}.

	\subsection{Contributions}
	\label{Contributions}
	
	The main contributions of this research can be summarized as follows: 
	
	\begin{itemize}
		\renewcommand{\labelitemi}{$\bullet$}
		\item Proposing a novel spatio-temporal stay extraction method to extract the stay regions of a mobile object by incorporating the introduces concepts of time-value and time-weighted centroid.
		\item Introducing a novel agglomerative hierarchical clustering method to merge the stay regions of a mobile object based on their geometries and constructing the geometries of the destinations of the mobile object.
		\item Developing a spatio-temporal partitioning method to partition the trajectory area of a mobile object into a grid with inhomogeneous cells containing the GOIs of the mobile object.
	\end{itemize}
	
	\subsection{Paper Organisation}
	\label{PaperOrganisation}
	
	The remaining part of the paper proceeds as follows: In section~\ref{RelatedWorks}, the related works focused on the partitioning of the trajectory area of mobile objects are discussed. In section~\ref{ProblemDefinition}, our problem is preliminarily defined. We introduce the concepts of time-value and time-weighted centroid in a GPS trajectory in section~\ref{ConceptOfTimeValueAndTimeWeightedCentroid}. In section~\ref{ExtractionOfStayRegions}, we present our proposed stay region extraction method and compare it with the related works. In section~\ref{FromStayRegionsToDestinationRegions}, a novel geometric similarity based agglomerative hierarchical clustering method for merging the similar stay regions and extracting the destination geometries is discussed and compared with the related works. In section~\ref{FromDestinationRegionsToGOIs}, our partitioning method which constructs a grid with inhomogeneous cells using the destination geometries constructed in the previous phase is introduced. In section~\ref{ComputationalComplexity}, we analyze the computational complexity of our method compared to the baselines. In section~\ref{ExperimentalResults}, the quality of our method compared to the related works is evaluated compared to the previous works. Finally, in section~\ref{ConclusionAndFutureWork}, the introduced method is summarized, and the achieved results and the future works are discussed.
	
	\section{Related Works}
	\label{RelatedWorks}
	
	In recent years, various works have considered trajectory data pre-processing, indexing, storage, and analysis~\citep{Zheng:TrajectoryComputingBook:2011}. These trajectories could be collected by social networks~\citep{LBSN:2011}, sensor networks~\citep{SensorPosition:2004}, RFIDs~\citep{RFID:2006}, WI-FI~\citep{MarkovOrderK:WLANPrediction:2006}, simulators~\citep{mousavi:mobisim}, internet of things~\citep{IOTIndoor:2014}, and cellular networks~\citep{CellularHMM:2010}. Among all of these kinds of trajectories, our work is focused on the trajectories collected by GPS sensors. GPS trajectories have been used in various research works in different applications~\citep{Zheng:TrajectoryComputingBook:2011}, however, our work is a pre-processing prerequisite for all of the applications which attempt to extract the GOIs of the mobile objects.

	\subsection{Partitioning Approaches}
	In the related works aiming to partition the trajectory area of a mobile object, five approaches have been taken. Following, we discuss the approaches and their capabilities and limitations. 
	
	\subsubsection{Grid with homogeneous cells}
	The first approach is to partition the trajectory area into a homogeneous grid to represent the regions of interest (e.g.~\citep{DestinationPrediction:2013}). The shape of the cells is often considered as triangular, square, rectangular, or hexagonal polygons. The main drawback of this approach is the degree of granularity of the cell. Coarse granularity leads to each of the grid cells cover a wide area which might include various POIs. The fine granularity results in the geometry of one POI to lie into different cells. These problems have significant drawbacks on the quality of the SVL extracted based on such grids.
	
	\subsubsection{Coverage Area Based}
	The second approach defines the POIs as the area being covered by a wireless accesspoint~\citep{MarkovOrderK:WLANPrediction:2006} in wireless networks or the area covered by base transceiver stations (BTS) of a cellular network~\citep{CellularHMM:2010}. The geometries of the POIs are constructed using circular area or hexagonal polygons around the access points or the BTS. The main problem with this approach is that estimating a fixed geometry for the area covered by a wireless access point or a BTS is not straight forward due to various reasons such as signal power, noise, and obstacles, particularly in the urban areas. Also, the problems, above-mentioned, related to the granularity of the grid cells remains. For example, the covered area by a BTS in a cellular network might cover a very wide area which includes various POIs, or a the covered area of an access point might not cover the whole area of a POI (covered by more than one access points).
	
	\subsubsection{Spatial Clustering Based}
	The third approach is to construct the geometries of the POIs based on the GPS track points in the trajectory datasets using simple spatial clustering methods without considering the temporal aspects of the GPS trajectories. Spatial clustering methods perform very similar to the classic clustering schemes such as KMeans~\citep{KmeansBasedClustering:2003}, Gaussian mixture model (GMM)~\citep{GMM:1993}, and DBSCAN~\citep{DJ-DBSCAN:2004}. These methods simply cluster the GPS points using measures such as the distance between GPS points or density connectivity in a two-dimensional Cartesian space, without taking the third dimension~\emph{time} into consideration, and partition the trajectory area based on the destination geometries constructed based on that clusters. 
	
	Another class of research works which can be categorized into spatial clustering based approach are research works~\citep{cecilia:nextplace2011,Li:movemine_journal2011} which have used \emph{frequency map based} spatial clustering methods for extracting significant places in the trajectory area. They partition the area into the very fine grid with equi-sized cells and assign a weight to each cell around each GPS point based on the duration of the GPS staying at that point. This weight assigned to each cell is computed based on the assumption that the real position of a mobile object has a normal distribution with standard deviation $\sigma=10m$~\citep{cecilia:nextplace2011}. Then they generate a frequency map which contains peaks that give information about the region of significant places. They consider regions that are above a predefined visit frequency threshold as POIs.     
	The main problem in the spatial clustering based partitioning approaches is the inaccuracy in the number and the geometries of the extracted POIs. They merely consider the density of the GPS track points in a neighborhood in the trajectory area as an indicator of a significant place or a POI. This assumption that the places which have more density of GPS track points are more significant for the user than the places with less density is not always true. Consider a mobile object often moves on a road network between its POIs regularly and repetitively. Obviously, during the journies between the POIs, there are some places which are being frequently visited and, therefore, have higher GPS track point density, while they are not the mobile objects POIs. For example, the conjunctions with traffic lights or the road segments with higher traffic loads often have a high density of GPS track points. These two approaches consider these kinds of places as POIs because they are not able to distinguish between POIs (with high density) and the non-POI places with nearly the same GPS point density. 
	
	\subsubsection{Speed Based}
	The fourth partitioning approach is taken by incorporating the speed restrictions in finding the stop and moves (E.g.~\citep{SpeedBasedPOIExtraction:2008,Bhattacharya:2012}). This approach assumes the clusters with the GPS track points with lower speed are more likely to be stop points. This approach is not applicable in GPS datasets where the GPS speed is not available, or the speed is not easily computable (e.g. in trajectories with low sampling rate or with large time gaps). Moreover, there are some scenarios where defining a threshold for maximum speed is not straight forward. For example, assume a mobile object carrying a GPS-enabled mobile phone. During the daily traveling activities, he might have different transportation modes (e.g. walk, bike, train, car, bus, etc.)~\citep{Zheng:TransportationMode:2010}. In each of the transportation modes the speed threshold should be different since the average walking speed is different to driving. Furthermore, even if we assume the same transportation mode for the mobile object throughout the trajectory (e.g. walk), places like shopping centers, zoos, parks, campuses, and so many other POIs exist where the mobile object stays in their geometry while keeping moving (speed is greater than zero).

	\subsubsection{Spatio-Temporal Clustering Based}
	The fifth approach~\citep{Zheng:LifePatterns:2009,Zheng:SimilarUsers:2010} employes \emph{time restricted} spatio-temporal clustering in extracting the stay regions and the destinations. They extract the stay regions based on predefined spatio-temporal restrictions. Then, they merge the stay regions to construct the destinations. 
	
	They define a valid stay region (a vicinity distance with radius $\Delta{D} \le D_{max}$) within which the mobile object has strayed (stopped or kept moving) for a time span $\Delta{T} \ge T_{min}$, where $T_{min}$, is a time span threshold. 
	The destinations which represent the POIs are extracted by clustering (merging) the stay centroid points of the extracted stay regions using density-based clustering methods such as OPTICS~\citep{Zheng:LifePatterns:2009}.
	This approach is highly used in research works such as~\citep{Zheng:LifePatterns:2009,Zheng:SimilarUsers:2010,Zheng:Geolife_1:2009,Zheng:Geolife_2:2008,Zheng:socialties:2014} conducted in Microsoft Research Asia. This approach effectively incorporates the temporal aspects of the mobile object trajectory into extracting the stay regions and as a result into the extracted destination regions. As a result, the places which the mobile object stays for a considerable time are selected, and the other places are filtered although they might have high point densities. 
	
	The research presented in~\citep{ProjectLachesis:2004} has a similar approach in the extraction of the stay regions with the difference that defines the time and vicinity distance based on the diameter of the extracted stay regions. The destinations are extracted based on the predefined maximum diameter of the destinations by merging the stay regions.  
	
	The fifth approach extracts the most meaningful and valid stay and destination regions because of their specific spatio-temporal definition of a valid stay region. Therefore, among all the works discussed above, in this paper, we choose the works discussed in the fifth approach as the baseline to compare the performance of our proposed partitioning method.

	\section{Problem Definition}
	\label{ProblemDefinition}        
	
	\begin{mydef}
		The trajectory of a moving object is a sequence of time stamped GPS observations (\emph{points}), $\mathcal{T}=\{p_1,p_2,...,p_n\}$, where $p_{i}=(t_{i},x_{i},y_{i})$
		indicates the spatio-temporal data of the moving object at time $t_{i}$. The parameters $t_{i}$, $x_{i}$, and $y_{i}$, are the time stamp and $(x,y)\in \mathcal{R}^2$ Cartesian plane of the moving object respectively. in our GPS trajectories, 
		$\forall p\in \mathcal{T},\quad i,j=1,2,\dotsc, \quad t^{p}_{i}>t^{p}_{j} \iff i>j$.
		There are no other guarantees such as constant sampling rate. 
	\end{mydef}
	
	\begin{mydef}
		The geometric similarity between a set of real GOIs $R=\{r_1,r_2,...,r_n\}$ and a set of estimated GOIs $G=\{g_1,g_2,...,g_m\}$ is defined as:    
		\begin{equation}
		\label{eq:GeometricSimilarity}
		GS(R,G)=\frac{1}{n}\sum\limits_{i}^{n}\sum\limits_{j}^{m}\frac{Area(r_i\cap g_j)}{Area(r_i \cup g_j)}.
		\end{equation}
		
	\end{mydef}
	
	Given a GPS trajectory $\mathcal{T}$, and a set of geometries of the real POIs $R=\{r_1,r_2,...,r_n\}$ covered by the MBR of $\mathcal{T}$, our objective is to propose the best (optimal) partitioning method to partition the MBR of the trajectory, $f_o: \mathcal{T} \rightarrow G$ which maximizes the geometric similarities between the real GOIs $R=\{r_1,r_2,...,r_n\}$ and their corresponding extracted GOIs $G=\{g_1,g_2,...,g_m\}$.    
	\begin{equation}
	%\nonumber
	f_o=\argmax_{f_i \in F} GS(f_i(\mathcal{T}),R),
	\end{equation}
	
	where $F=\{f_1,f_2,...,f_k\}$ is the set of different partitioning methods, e.g., $\{$diameter based, density based, geometric similarity based$\}$.
	
	Subject to     
	\begin{itemize}
		\renewcommand{\labelitemi}{$\checkmark$}
		\item $\forall g_j$ and $g_k \in G: \:$ if $j \neq k$ then $Area(g_j \cap g_k) = 0$,    
		\item $\forall p_t \in \mathcal{T}, \exists g_j \in G \; \mid \; p_t \cap g_j \neq \O$.
	\end{itemize}
	
	The first constraint guarantees that the geometries of extracted partitions are mutually disjoint. The second constraint ensures that all the GPS points in the trajectory can be assigned to one and only one partition. 
	
	\section{Concepts of Time-Value and Time-Weighted Centroid in Trajectory Data}
	\label{ConceptOfTimeValueAndTimeWeightedCentroid}
	
	In this section, we briefly introduce the concepts of time-value and time-weighted centroid which will be used in the first phase of our partitioning method to improve the performance of the stay region extraction.
	
	In the process of collecting GPS observations which are often done by GPS devices installed on vehicles or mobile phones, ideally, we would like to collect each GPS observation with constant sampling rate. For example, we would like to have one sample point every 10 seconds or every one minute. However, due to various reasons, it is not always applicable. GPS sensors installed on mobile phones consume a considerable amount of power. So people usually tend to keep their GPS sensor off. This fact has a dramatic impact on the quality of the collected GPS trajectories. Another reason is poor GPS coverage in places such as urban environments and particularly in indoor locations. Besides, the process of GPS data collection is often terminated by the user for long periods (e.g. in the car parks). 
	
	The time gap between two consecutive GPS observations can be short or considerably long. The long time gaps often take place when a vehicle is parked at a car park,  or a mobile device is switched off. We consider the time gap between two consecutive GPS points in a trajectory as a significant influencing factor.
	
	\begin{mydef}
		For a GPS point $p_{i}$ in the trajectory $\mathcal{T}$, we define the time-value as: 
		\begin{equation}
		\label{TimeValue}
		tv^{p}_{i}=t^{p}_{i+1}-t^{p}_{i} \quad i=0,1,...,n.
		\end{equation} where $t^{p}_{i}$ indicates the time stamp of point $p_i$. 
	\end{mydef}
	
	As an example, we can consider the simple problem of computing the centroid of a set of GPS points to address the effectiveness of considering the time-value of GPS points in trajectory data processing.
	
	\begin{mydef}
		The centroid $c_i=(c^{x}_{i},c^{y}_{i})$ of a set of points $PS=\{p_{m},p_{m+1},...,p_{n}\}$, is often computed as: 
		\begin{equation}
		\label{ComputeCentroid}
		c^{x}_{i}=\frac{\sum_{i=m}^{n} x^{p}_{i}}{|PS|},\quad
		c^{y}_{i}=\frac{\sum_{i=m}^{n} y^{p}_{i}}{|PS|}
		\end{equation}
		where, $x^{p}_{i}$ and $y^{p}_{i}$ are the $x$ and $y$ coordinates of point $p_i$, and $|PS|$ is the cardinality of the point set $PS$.
	\end{mydef} 
	
	In Eq.~\ref{ComputeCentroid}, the values of all the GPS points are considered the same in computing the centroid. Contrary to the previous works, we incorporate the time-value of each GPS point $tv^p_i$ in computing the centroid resulting in the time-weighted centroid of the set of GPS points.
	
	\begin{mydef}
		The time-weighted centroid ($twc^{x}_{i},twc^{y}_{i}$) of a set of points $PS=\{p_{m},p_{m+1},...,p_{n}\}$ is defined as: 
		\begin{equation}
		\label{ComputeTimeWeightedCentroid}
		twc^{x}_{i}=\frac{\sum_{i=m}^{n} x^{p}_{i} \times tv^{p}_{i}}{\sum_{i=m}^{n} tv^{p}_{i} \times |PS|}\quad
		twc^{y}_{i}=\frac{\sum_{i=m}^{n} y^{p}_{i} \times tv^{p}_{i}}{\sum_{i=m}^{n} tv^{p}_{i} \times |PS|}\quad
		\end{equation}  
		where, $tv^{p}_{i}$ is the time-value of point $p_i$ computed using Eg.~\ref{TimeValue}, and $|PS|$ is the cardinality of the point set $PS$.
	\end{mydef}
	
	In Eq.~\ref{ComputeTimeWeightedCentroid}, the time-value of each GPS point is considered as the weight or degree of significance of each point in computing the centroid. By this, we discriminate our GPS points based on the value of information they give us about the location of the mobile object. By incorporating the time-value, the centroid will be more biased to and closer to the locations where long term stops have taken place.

	\section{Methodology}
	\label{Methodology}
	
	Our partitioning method has three phases. (i) Spatio-temporal extraction of stay regions, (ii) Constructing the destination regions based on the extracted stay regions, (iii) Partitioning the MBR of the trajectory based on the extracted destinations. Following, we discuss each phase of the method and compare them with the related works in detail.
	
	\subsection{Spatio-Temporal Extraction of Stay Regions}
	\label{ExtractionOfStayRegions}
	
	Extraction of the stay regions of a mobile object is the first phase of our spatio-temporal partitioning method. Aiming for that as the first step We convert the GPS trajectory into a sequence of \emph{stays} and \emph{moves}. We define a stay $s_i$ as an event which has been taken place within the trajectory period of a mobile object. The event has happened in a geometric region or neighborhood called a stay region. A stay region is an area in which the mobile object spends some time $\Delta{t} \geq T_{min}$. During this time, the mobile object can be either moving or stopping provided that it does not pass the boundary of the region. The boundary of the region is calculated based on the roaming distance $\Delta{d}\leq D_{max}$ which is the maximum distance that a moving object can stray from the centroid of the stay region. For example, if a vehicle has stopped in a car park for 8 hours starting from 9 AM to 5 PM, the event is the visit to the car park, the starting time of the event is 9 AM (arrival time) and the ending time of the event is 5 PM (departure time). Each stay has a set of GPS points (point set) which indicates the GPS observations which were collected within the stay period. 
	
	\begin{mydef}
		We define stay $s_{i}, \: i=1,2,...n$ as $s_{i}=(id_{i},g_{i},ps_{i},c_{i},at_{i},dt_{i})$,
		where $id_{i},ps_{i},g_{i},c_{i},at_{i},dt_{i}$ are the identification, geometry, point set, centroid, arrival time and departure time of of $s_{i}$, respectively. $ps_i$ is a sub trajectory of the mobile object trajectory which is defined as a set of consecutive points $\{ p_{m},p_{m+1},...,p_{n}\}$, where $\forall k, m<k\le n$, $Dist(c_i,p_{k})\le D_{max}$, and $Dist(c_i,p_{n+1})>D_{max}$. 
		The parameter $c_i$ referes to the centroid of the points in $ps_i$ and $g_{i}$ is the the convex polygon of the point set $ps_i$. 
	\end{mydef}

	\IncMargin{1em}
	\begin{algorithm}[t!]
		\label{alg:TimeWeightedCentroidBasedStayExtraction}   
		\SetKwInOut{Input}{input}\SetKwInOut{Output}{output}
		\Input{$P$ (A set of GPS points),
			vicinity distance threshold $D_{max}$,  
			time span threshold $T_{min}$}
		\Output{A set of Stays $S$ where $s_{k}=(id_{k},g_{k},ps_{k},c_{k},at_{k},dt_{k})$} 
		\KwData{Coordinate twc}
		\BlankLine
		\Indp  
		$\Delta{d} \gets 0$,
		$\Delta{t} \gets 0$,
		$i \gets 0$,
		$j \gets 0$,
		$token \gets 0$ \\ 
		\ForEach{$p_i \in P$}
		{$p_i.tv \gets$ ComputeTimeValue($p_i$)}    
		\While{$i<|P|$}{
			$ps_k$.insert($p_i$) \\
			$twc$ $\gets$ TimeWeigtedCentroid($ps_k$)\\ 
			$j \gets i+1$ \\
			$token \gets 0$ \\     
			\While{$j<|P|$}{
				$\Delta{d} \gets$ EucDistance($twc,p_j$)\\  
				\If{$\Delta{d}>D_{max}$}{
					$\Delta{t} \gets (t^{p}_{j}+tv^{p}_{j})-t^{p}_{i}$ \\         
					\If{$\Delta{t} \ge T_{min}$}{
						$id_k \gets k$ \\
						$g_k \gets$ ComputeConvexHull($PS$) \\ %\tcp*{Geometry of the computed stay $s_k$}
						$c_k \gets$ GeometryCentroid($g_k$) \\ %\tcp*{Centroid of the $g_k$}
						$at_k \gets t^{p}_{i}$ \\ % \tcp*{Arrival time of the computed stay $s_k$}
						$dt_k \gets t^{p}_{j}+tv^{p}_{j}$ \\ % \tcp*{Departure time of the computed stay $s_k$}
						$s_{k}=(id_{k},g_{k},ps_{k},c_{k},at_{k},dt_{k})$ \\
						$S$.insert($s_k$) \\
						$i \gets j$  \\
						$token \gets 1$\\ 
						$k \gets k+1$  \\                  
						break
					}        
				}
				$ps_k$.insert($p_j$) \\ 
				twc $\gets$ TimeWeightedCentroid($ps_k$) \\
				$j \gets j+1$  \\           
			}
			\If{$token \ne 1$}{$i \gets i+1$}
		}
		\Return{$S$} 
		\caption{Time-Weighted Centroid Based Stay Region Extraction (Our Method)}
	\end{algorithm}
	\DecMargin{1em}

	The definition of a $\Delta{d}$ in our approach is different to the previous work~\citep{ProjectLachesis:2004} and ~\citep{Zheng:LifePatterns:2009}. In~\citep{ProjectLachesis:2004}, $\Delta{d}$ is defined as the Euclidean~\emph{diameter} of the coordinates of elements of a stay. This means that the diameter (the longest Euclidean distance between two points in the set) of a stay must not be greater than $\Delta{d}$. They iteratively add a point to the sub-trajectory $p_i$ and recompute the diameter of $p_i$. If the diameter, remains less than $D_{max}$ after adding the new point, they keep the point in $p_i$. Otherwise, they remove the point from $p_i$, store $s_i$, and start constructing a new stay. Ye et al., in~\citep{Zheng:LifePatterns:2009} have taken the very first chronological point in each stay as the reference point and have defined $\Delta{d}$ as the Euclidian distance between each new point and the reference point. They do not refresh the reference point coordinate when adding a new point to the stay. While in our method, $\Delta{d}$ is defined as the Euclidian distance between the time-weighted centroid of the stay (reference point) to the new point which is being examined (line 11 in Alg.~\ref{alg:TimeWeightedCentroidBasedStayExtraction}). In other words, we use the time-weighted centroid of the point set in the current stay as the reference points instead of the very first point of the stay. 
	
	The calculation of the parameter $\Delta{t}$ in our method is different to the previous works as well. In the previous works, the parameter is defined as $\Delta{t} \gets (t^{p}_{j})-t^{p}_{i}$, which means the time difference between the first and the last GPS observation in the stay. In our method we incorporate the time-value of the last point ($tv^{p}_{j}$) and define the $\Delta{t}$ as: $\Delta{t} \gets (t^{p}_{j}+tv^{p}_{j})-t^{p}_{i}$. This means that we consider the time gap between the last point and its successor point in the trajectory to compute $\Delta{t}$.  
	
	In our method (presented in Alg.~\ref{alg:TimeWeightedCentroidBasedStayExtraction}), having a current stay, for each new GPS observation in the trajectory, if the condition $\Delta{d} > D_{max}$ and $\Delta{t} > T_{min}$, are true, we close the current stay, store it, and make a new stay with the GPS observation as the first point in its point set. Otherwise, we add the new GPS observation to the point set of the current stay, update the time-weighted centroid of the current stay, and keep examining the next points in the trajectory. In other words by adding each point to the stay, we refresh the coordinate of the reference point as the time-weighted centroid of the points of the current stay. We assign a unique numeric identification $id_i$ to each stay $s_i$. The parameter $at_i$ indicates the arrival time of stay $s_{i}$ which is the time that the moving object has arrived in the region ($g_i$). Similarly, $dt_i$ is the departure time of stay $s_{i}$. 
	
	To calculate the geometry of the region $g_i$ within which the stay $s_{i}$ has taken place, we compute the convex hull of the set of points $ps_i=\{p_{1},p_{2},...,p_{n}\}$. The convex hull of a set of points is the smallest polygon that contains all of the points~\citep{ConvexHull:1979}. Then, we add a predefined geometric buffer around the convex hull polygon to compensate for the GPS noise. The width of the buffer is set to $10$ meters~\citep{GPSStandards:2008}. Fig.~\ref{fig:StayRegions} shows the extracted stay regions based on the trajectory shown in Fig.~\ref{fig:Trajectory}. 
	
	\subsection{From Stay Regions to Destination Regions}
	\label{FromStayRegionsToDestinationRegions}
	
	Assume a moving object visits a certain place every day (e.g. home). If we extract every stay throughout a trajectory period with a long time duration, e.g., one year or more, we would have at least 365 extracted stays with approximately the same geometry. At this stage, we need to merge the duplicated stays which represent the same destination region (e.g. car park). Aiming for that, we detect the stays that have approximately the same geometry and merge (cluster) them together resulting in a set of \emph{destinations} with unique geometries and Identification. 
	
	%Note that the geometry of the stay is not necessarily the same as the geometry of the visited place because people might park their cars near their house and not always at their house. Moreover, after entering the home car park (indoor location) the GPS coverage might be lost, or the GPS device might be switched off. So, we cannot expect the stay extractor methods to estimate the geometry of the destinations themselves. 
	
	In related works, there are two major schemes for merging the stay regions. They cluster the stay points (the centroids of the GPS points stored in each stay) such that the stay points that have close distance are clustered into the same destination. Ye et al. in~\citep{Zheng:LifePatterns:2009}, have used the density of neighborhood of stay points in a group as a measure of similarity in the clustering process. In other words, the stay points which have more dense neighbors (stay points with many nearby neighbors closely packed together) make a cluster (destination) and stay points that lie alone in low-density regions are considered as outliers. The result of running OPTICS~\citep{OPTICS:1999} clustering algorithm on the set of stay points is a set of destinations (each destination is a set of stay points). 
	
	One major problem of density-based clustering methods such as OPTICS is that it is required to define two parameters neighborhood distance $\epsilon$ $eps$ (neighborhood distance) and the $(minPts)$ (minimum number of points required to form a dense region). The performance and output of the methods are strongly sensitive to the values chosen for these two parameters. For instance, in our application, if we choose a relatively big value for parameter $(minPts)$, the density-based clustering methods will consider lots of stay points as noise or outliers in the clustering and eliminate them because they have the lower density of neighbors than $(minPts)$. 
	
	Hariharan et al., in~\citep{ProjectLachesis:2004} have used a clustering method which finds each pair of stay points which have maximum similarity to each other and merge them together iteratively. They have defined a similarity criterion which indicates if the diameter of the resulting region of merging two stay points is less than or equal to a given threshold $D_{min}$, these two stay points will be merged. This process continues until all similar stay points are merged. 
	
	\begin{figure*}[t!]
		\centering
		\subfigure[t][GPS Trajectory]{\label{fig:Trajectory}\includegraphics[width=0.19\linewidth]{Trajectory.png}}
		%~\hspace{1em}
		\subfigure[t][Stay Regions]{\label{fig:StayRegions}\includegraphics[width=0.18\linewidth]{StayRegions.png}} %~\hspace{1em}
		\subfigure[t][$J_{min}=0$]{\label{fig:Destination_0}\includegraphics[width=0.18\linewidth]{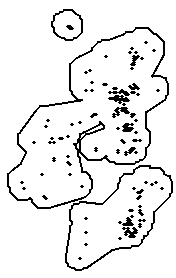}}
		%~\hspace{1em}
		\subfigure[t][$J_{min}=0.10$]{\label{fig:Destination_01}\includegraphics[width=0.18\linewidth]{Destination_01.png}}
		%~\hspace{1em}
		\subfigure[t][$J_{min}=0.20$]{\label{fig:Destination_02}\includegraphics[width=0.18\linewidth]{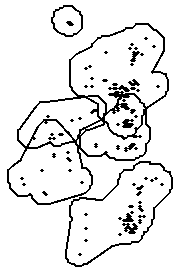}}
		\centering
		\caption{Results of the Destination Extraction Phase With Different Values of $J_{min}$.}
		\label{fig:SimilarityImpactOnMerging}
	\end{figure*}

	\IncMargin{1em}
	\begin{algorithm}[]
		\label{alg:GeometricSimilarityBasedDestinationExtraction}
		\SetKwInOut{Input}{input}\SetKwInOut{Output}{output}
		\Input
		{A set of stay regions $S$,
			Jaccard similarity threshold $J_{min}$
			visit frequency threshold $F_{min}$ 
		}
		\Output{A set of destination regions $D$\\
		} 
		\KwData{Destination $d$,
			$InterList$,
			$RTreeIndex$
		} 
		\BlankLine
		\Indp        
		$JSim_{max} \gets 0$,
		$JSim \gets 0$,
		$firstSimIndex \gets 0$,
		$secondSimIndex \gets 0$\\
		\ForEach{$s_i \in S$}
		{$f_i \gets 1$
		}
		\While{$(JSim_{max}<J_{min})$}{
			RTreeIndex.Update($S$)\\
			$i \gets 0$,
			$JSim_{max} \gets 0$  \\
			\While{$(i<|S|)$}
			{$interList \gets $ RTreeIndex.FindIntersectingStays($s_i$)  \\
				$j \gets 0$  \\
				$JSim \gets 0$\\
				$JSim_{max} \gets 0$  \\
				\While{$(j<|interList|)$}
				{
					$JSim \gets \sfrac{Area(g_i \bigcap interList[j])}{Area(g_i \bigcup interList[j])}$  \\          
					\If{$(JSim>JSim_{max})$}{$JSim_{max} \gets JSim$  \\  
						$maxIndex \gets FindIndex(S,interList[j])$ %\tcp*{finds the index of $interList[j]$ in $S$}
					}
					$j \gets j+1$ 
				}                
				$i \gets i+1$
			}
			\If{$(JSim_{max}>J_{min})$}
			{$s_i.MergePoints(s_{maxIndex})$ \\ %\tcp*{Merges GPS Points} 
				$g_i \gets g_i \cup g_{maxIndex}$ \\%\tcp*{Merges geometries}
				$f_i \gets f_i+f_{maxIndex}$ \\ %\tcp*{Computes Visit Frequencies}
				$S.remove(s_{maxIndex})$ \\%\tcp*{Removes $s_{maxIndex}$ from $S$ after being merged to another stay}
			}\Else{break}    
		}
		\ForEach{($s_i \in S)$}
		{$D.insert(s_i)$
		}
		\ForEach{($d_k \in D)$}
		{
			\If{$f_k < F_{min}$}
			{$D.remove(d_i)$
			}
		}
		\Return{$D$}    
		\caption{Geometric Similarity Based Destination Detection Method}
	\end{algorithm}
	\DecMargin{1em}
	
	In our method, we incorporate the geometries the stay regions in extracting the destinations instead of only considering the density or distance of the stay points. In the process of clustering, we define a criterion that helps us control our merging process in our hierarchical clustering method.
	
	Alg.~\ref{alg:GeometricSimilarityBasedDestinationExtraction} presents the pseudo-code of our method. We use R-Tree indexing method to index the geometry of each stay $g_i \in S$. Subsequently, at each step, we send a query to the R-Tree to find only the clusters which their geometry intersects with the current stay geometry. The result is a list of stays ($interList$). Then, we compute the most similar stay geometry in $interList$ to our current stay region ($g_i$). After finding the similarity of all pairs in $S$, if $JSim_{max}=0$, this means that there is no intersecting pair of stay regions in $S$. If $JSim_{max}>0$, then there is still, at least, a pair of interesting stay regions in $S$. $s_{maxIndex}$ represents the stay region that has the highest similarity to our current stay $s_i$ in the second loop in Alg.~\ref{alg:GeometricSimilarityBasedDestinationExtraction}. To decide whether we need to merge the current stay $s_i$ and $s_{maxIndex}$, we compare their similarity coefficient ($JSim_{max}$) with $J_{min}$. If ($JSim_{max}>J_{min}$), we merge the two stays by adding all the GPS points in stay $s_{maxIndex}$ to $s_i$ and computing the new region geometry of our current stay $g_i$ as the geometric union of two geometries. 
	
	After merging stay region pairs, we compute the visit frequency of the resulting stay region as the sum of visit frequency of the current stay $f_{i}$ and the most similar stay region $s_{maxIndex}$. We consider the frequency of visits to each destination as a useful criterion for selecting the significant destinations (POIs). As a result, we can decide whether we consider a cluster of stay regions (destination) as a POI or consider it as a trivial cluster (noise). The last loop in Alg.~\ref{alg:GeometricSimilarityBasedDestinationExtraction} removes the destinations which have been visited with the frequency less than $F_{min}$.
	
	We define the similarity of two geometries $(g_i,g_j)$ as the measure of similarity of two stay regions as follows:        
	\begin{equation}
	\label{eq:JaccardSimilarity}
	GS(g_i,g_j)= \frac{Area(g_i\cap g_j)}{Area(g_i \cup g_j)}.
	\end{equation}    
	
	Fig~\ref{fig:SimilarityImpactOnMerging} shows the destinations extracted from the a set of stay regions using different values for parameter $J_{min}$. The higher value of $J_{min}$ leads to higher number of overlapping destination regions being extracted. Interestingly, if we set $J_{min}=1$, we will have exactly all the stay regions extracted as destination regions because the probability that two stay regions have identical convex hulls (and accordingly $JSim=1$) is approximately zero. 
	
	\subsection{From Destination Regions to Geometries of Interest (GOIs)}
	\label{FromDestinationRegionsToGOIs}
	
	\begin{figure*}
		\centering
		\subfigure[t][Micro-Grid]{\label{fig:Micro-Grid}\includegraphics[width=0.20\linewidth]{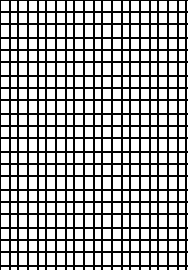}}
		\subfigure[t][Destination-Grid]{\label{fig:Destination-Grid_01_PCS}\includegraphics[width=0.19\linewidth]{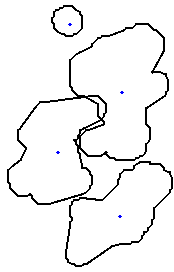}}
		\subfigure[t][GOI-Grid (PCS)]{\label{fig:Goi_01_PCS}\includegraphics[width=0.19\linewidth]{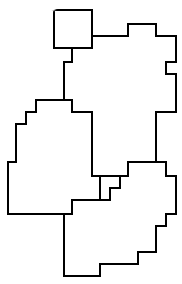}}
		\subfigure[t][GOI-Grid (GS)]{\label{fig:Goi_01_JS}\includegraphics[width=0.19\linewidth]{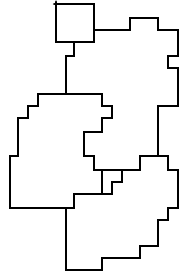}}
		\subfigure[t][Final-Grid (GS)]{\label{fig:Final_01_JS}\includegraphics[width=0.19\linewidth]{FinalGrid.png}}
		\centering
		\caption{Partitioning Results Using Two Geometric Similarity Metrics.}
		\label{fig:GeometricSimilarity}
	\end{figure*}
	
	Having extracted the destination regions and their estimated geometries, as the final phase of our partitioning method, based on the geometries of the destination regions, we partition our trajectory area into a grid area with inhomogeneous cells such that both characteristics of a valid partitioning (discussed in section~\ref{ProblemDefinition}) are guaranteed. 
	
	Firstly, we make a grid called micro-grid $MG$ with equi-sized rectangular shaped cells with very fine granularity. The grid covers the area minimum bounding rectangle (MBR) of our GPS trajectory $\mathcal{T}$. Also we make a grid composed of geometries of all destination regions. We refer to this grid as destination-grid ($DG$). Then we convert the destination-grid ($DG$) to a grid called \emph{GOI-Grid} ($GG$) with mutualy disjoint cells. Aiming for that, for each cell $m_i \in MG$, we find the cell in $d_j \in DG$ which maximizes the geometric similarity (Eq.~\ref{eq:JaccardSimilarity}) with $m_i$.     
	\begin{equation}
	\nonumber
	d_j \in DG= \argmax_{d_j} GS(c_i,d_j).
	\end{equation}
	
	We also defined and examined an alternative similarity metric as the Euclidean distance between the centroid of the polygon of cell $i$ in micro-grid ($c^m_i$) to the centroid of the polygon of cell $j$ in destination-grid ($c^d_j$). We call this similarity as polygon centroid similarity (PCS).    
	\begin{equation}
	\nonumber
	\label{PCS}
	PCS(m_i,d_j)=\frac{1}{EucDistance(c^m_i,c^d_j)}.
	\end{equation}
	
	Next, we label $m_i \in MG$ as a cell that represents a tiny part of the destination $d_j$ in the destination-grid. We continue this process until there are no remaining unlabeled cells in $MG$ which have an intersection with any of the cells in $DG$. By merging the geometries of all the cells in $MG$ labeled with ID of each cell in $DG$, we make the GOI-Grid (depicted in Fig~\ref{fig:Goi_01_JS}). 
	
	Since we find the most similar destination for each cell $m_i \in MG$ and label it to be a part of the geometry of only one destination, all GOIs in the GOI-Grid are mutually disjoint, the first condition of a valid spatial partition is guaranteed. However, the grid does not cover all of the area of the MBR of the trajectory. This means that there might be a GPS observation that does not lie in the geometry of one of the GOI-Grid cells. To tackle this problem, we insert all the cells $m_i \in MG$, which were not already been labeled, into the GOI-Grid. The resulting grid in referred to as final-grid (Fig.~\ref{fig:Final_01_JS}) which is composed of the GOIs of the mobile object, and the tiny cells with unique IDs. As a result, all of the area of the MBR of the trajectory $\mathcal{T}$ are covered by either a GOI or a tiny cell in final-grid. 
	
	Fig.~\ref{fig:Goi_01_JS} and Fig.~\ref{fig:Goi_01_PCS} show the performance of both similarity metrics (GS and PCS) in the partitioning. As is seen, the resulting GOI-Grid using geometric similarity resembles the Destination-Grid much better because the PCS based method is biased to the centroid of the polygon and makes the shape of the resulting polygon less similar to the corresponding cell in the destination-grid.

	\section{Computational Complexity}
	\label{ComputationalComplexity}
	
	In the stay extraction phase, in the worst case, the time complexity of our method is $\mathcal{O}(n^2)$ for $n$ track points in the trajectory. However, in practice, since the sum of the track point of the extracted stays are considerably fewer than $n$. Note that, the inner loop in Alg.~\ref{alg:TimeWeightedCentroidBasedStayExtraction} deals with computing the centroids of the stay regions which depends on the number of the track points in each stay. Since a large number of track points in the trajectory are not clustered in the stays (due to restrictions of a valid stay), the sum of the track points of the stays is much lower than $n$. Computing the diameter of the stays in the work proposed by~\citep{ProjectLachesis:2004} is more complex than computing the centroid in our method, since for computing the diameter, we need to compute the distance of each point to all of the other points in the cluster with the time complexity of $\mathcal{O}(n^2)$. Therefore, we can consider the complexity of the method, in the worst case, $\mathcal{O}(n^3)$ for $n$ track points in the trajectory. The time complexity of the method proposed by~\citep{Zheng:LifePatterns:2009} is $\mathcal{O}(n)$ which is lower than our method since they do not refresh the coordinate of the reference point while making the stay region. 
	
	In the destination extraction phase, Zheng et al., in~\citep{Zheng:LifePatterns:2009} have used OPTICS~\citep{OPTICS:1999} clustering method. The time complexity of OPTICS algorithm is $\mathcal{O}(n^2)$. The time complexity of the hierarchical clustering method provided by Hariharan et al.~\citep{ProjectLachesis:2004} is $\mathcal{O}(n^3)$ in the worst case. The complexity of our method (Alg.~\ref{alg:GeometricSimilarityBasedDestinationExtraction}) in the worst case, is $\mathcal{O}(n^3)$. We use R-Tree indexing method to reduce the runtime of our method in finding the intersecting cells. The most costly part in our method is finding the degree of similarity between two geometries (Eq.~\ref{eq:JaccardSimilarity}).
	
	In the partitioning phase, in the process of assigning each of the cells in micro-grid to the cells in GOI-Grid, in the worst case, the complexity of the method is $\mathcal{O}(nm)$, where $n$ is the number of cells in the $MG$ and $m$ is the number of cells in the destination-grid. Therefore, the granularity of $MG$ has a significant impact on the runtime of our method. To increase the efficiency of the method we use R-Tree indexing to index the cells in the destination-grid. 
	
	\section{Experimental Results}
	\label{ExperimentalResults}
	
	\begin{table*}[t!]
		\centering
		\begin{tabular}{ccc} 
			%\centering
			%\toprule
			\cmidrule(r){1-3}
			\centering
			Stay Extraction Method & Number of Extracted Stay Regions & Number of Single Sized Stay Regions \\
			\midrule
			Reference Point Based~\citep{Zheng:LifePatterns:2009} & 3568 & 0 \\
			Diameter Based~\citep{ProjectLachesis:2004} & 3587 & 0 \\
			Time-Weighted Centroid Based (Our Method) & 4127 & 292\\
			\bottomrule
		\end{tabular}
		\caption{Spatio-temporal Stay Region Extraction Results ($T_{min}=60min$)}
		\label{table:StayExtractionResults60min}   
	\end{table*}
	
	In this section, we analyze the performance of our proposed method in comparison with the baselines. In our evaluations, we use a dataset of GPS trajectories collected in Anchorage, Alaska, USA as a part of the project FreeSim~\citep{FreeSim:2007,FreeSim:2009}. The trajectory we use in this paper has been collected from a vehicle for the duration of about 42 months from 2010 to 2013 with varying sampling rate from one sample every 10 seconds to one sample every two minutes. We used ELKI machine learning library~\citep{ELKI:2015}, to implement the OPTICS clustering algorithm. Following we discuss the results in each of the three phases.
	
	\subsection{Stay Extraction Experimental Results}
	\label{StaysExtractionExperimentalResults}
	
	We implemented the method presented in Alg.~\ref{alg:TimeWeightedCentroidBasedStayExtraction} and two methods proposed in~\citep{ProjectLachesis:2004}, and~\citep{Zheng:LifePatterns:2009}. Table~\ref{table:StayExtractionResults60min} presents a comparison of the experimental results for each stay extraction method. As it is seen, our time-weighted clustering method outperforms the other two methods in the number of extracted clusters. Moreover, our method detects and report the stay regions with the single point while the other methods simply lose the stay regions.
	
	\begin{figure*}[t!]
		\centering
		\subfigure[t][Map of the Selected Area in Anchorage, Alaska, USA (Bing) ]{\label{fig:MapOfTheSelectedArea_Bing}\includegraphics[width=1\linewidth]{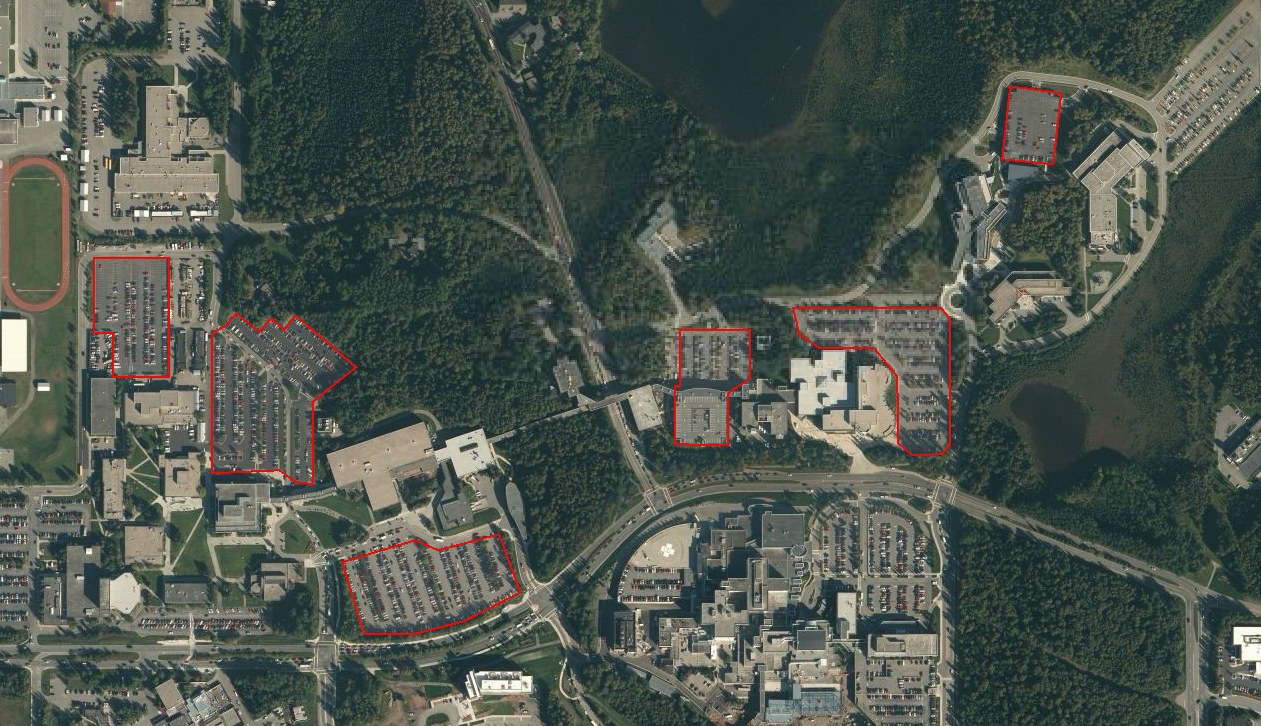}}
		\subfigure[t][Map of the Selected Area in Anchorage, Alaska, USA (Mapnik) ]{\label{fig:MapOfTheSelectedArea_Mapnik}\includegraphics[width=0.48\linewidth]{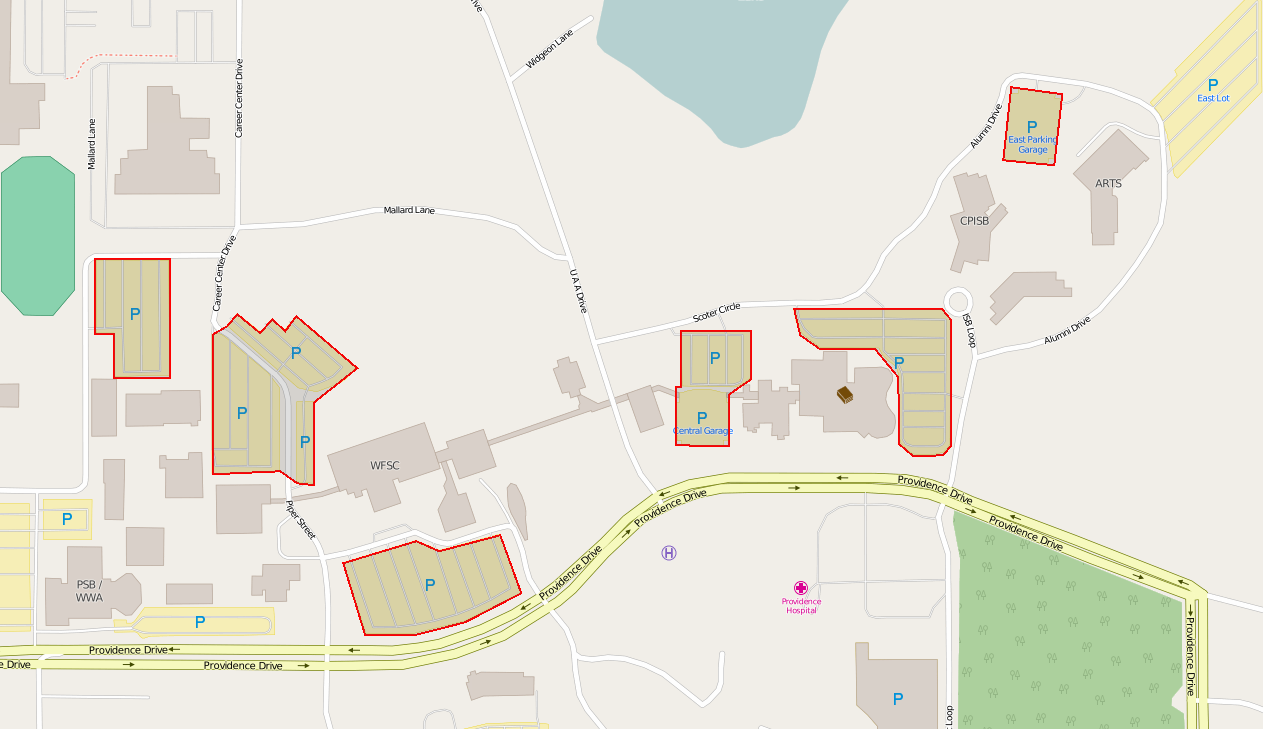}}
		~\hspace{1em}
		\subfigure[t][Diameter Based Stay Extraction Method ($T_{min}=60min, Diam_{max}=200m$)]{\label{fig:DiameterBasedStays}\includegraphics[width=0.48\linewidth]{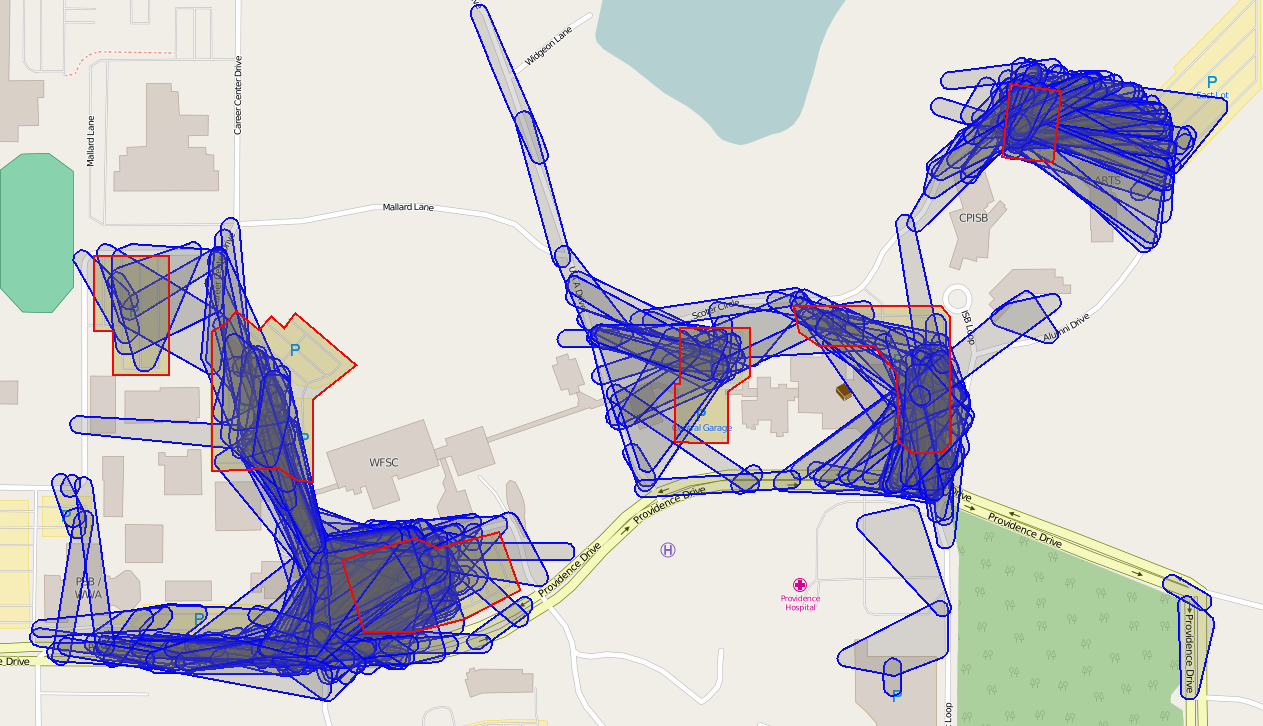}}
		\subfigure[t][Reference Point Based Based Stay Extraction Method ($T_{min}=60min, Diam_{max}=200m$)]{\label{fig:ReferencePointBasedStays}\includegraphics[width=0.48\linewidth]{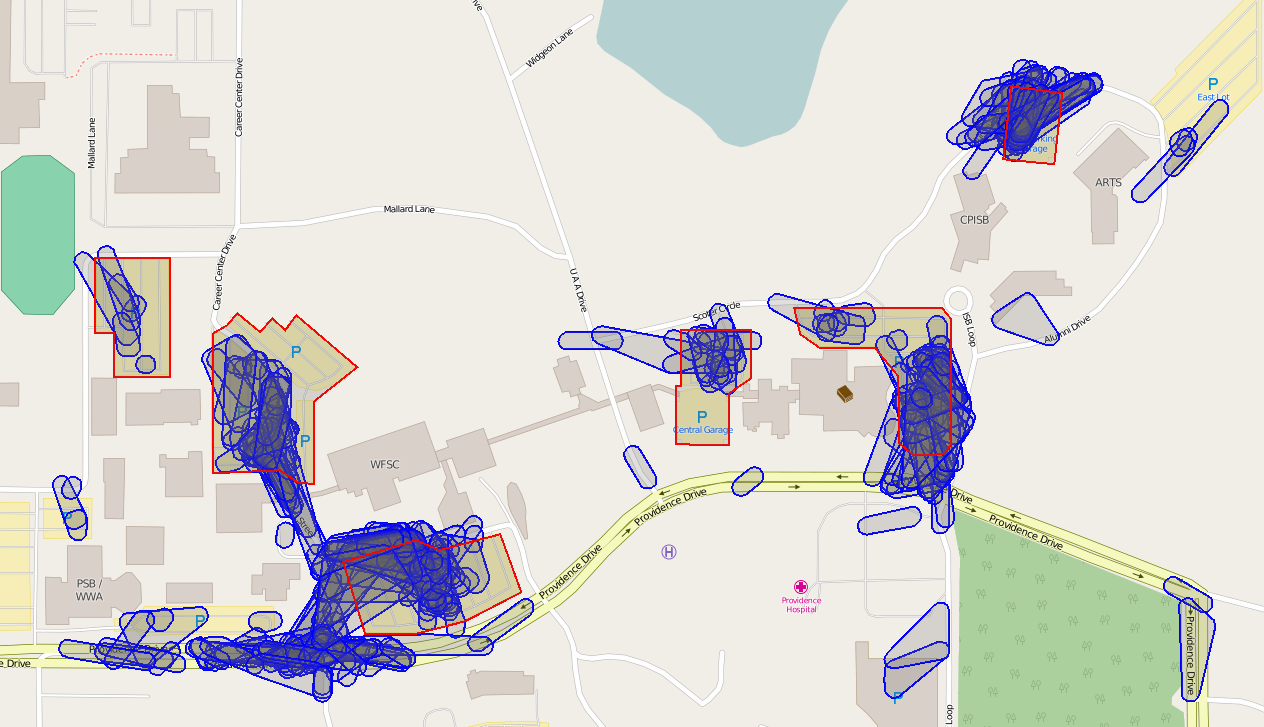}}
		~\hspace{1em}
		\subfigure[t][Time-Weighted Centroid Based Stay Extraction Method ($T_{min}=60min, D_{max}=100m$)]{\label{fig:TimeWeightedCentroidBasedStays}\includegraphics[width=0.48\linewidth]{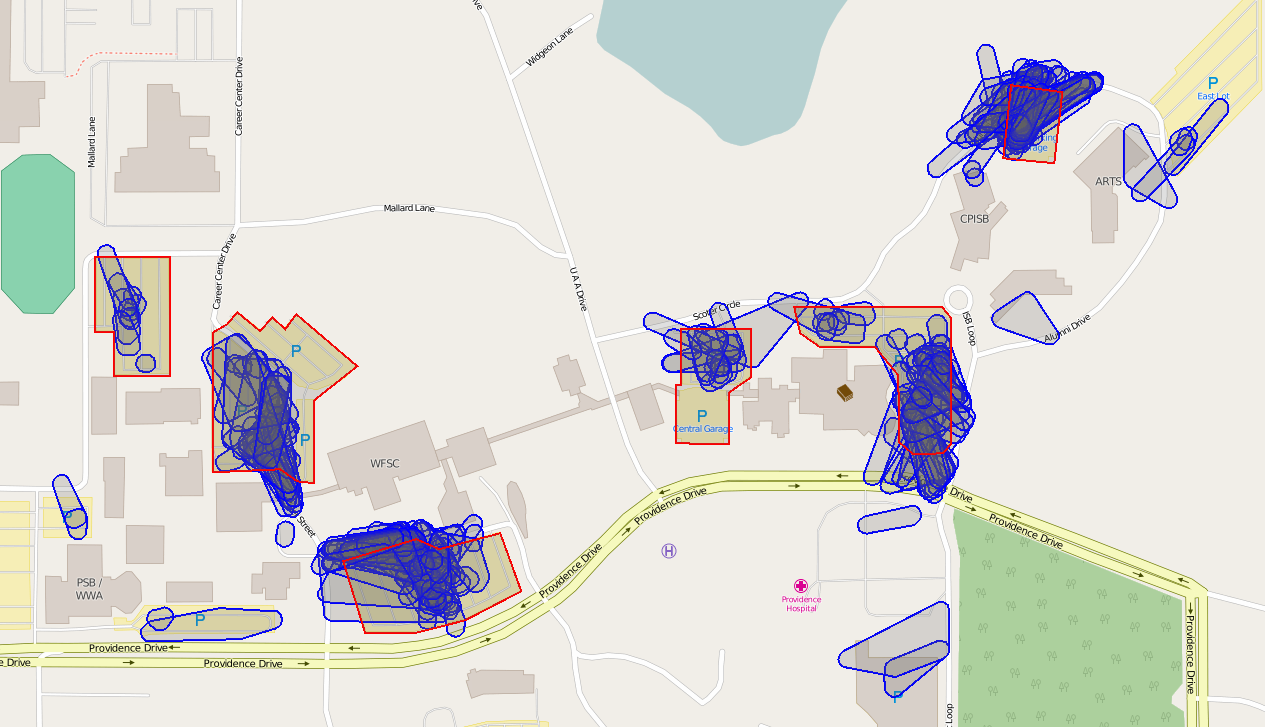}}
		\caption{Stay Region Extraction Results}
		\label{fig:StayRegionExtractionResults}
	\end{figure*}

	\begin{table*}[]
		\centering
		\begin{tabular}{cccc} 
			%\toprule
			\cmidrule(r){1-4}
			Destination Extraction Method & Parameters & Number of Stays & Number of Destinations \\
			\midrule
			Diameter Based~\citep{ProjectLachesis:2004} & $Diameter_{min}=200m$ & 3587 & 304\\
			Diameter Based & $Diameter_{min}=300m$ & 3587 & 166\\
			Diameter Based & $Diameter_{min}=400m$ & 3587 & 128\\
			%Diameter Based & $Diameter_{min}=500m$ & 3587 & 92\\ 
			\\
			Density Based~\citep{Zheng:LifePatterns:2009} & $eps=100m,minPts=3$ & 3566 & 456\\
			Density Based & $eps=100m,minPts=6$ & 3566 & 206\\
			Density Based & $eps=100m,minPts=9$ & 3566 & 120\\
			%Density Based & $eps=100m,minPts=12$ & 3566 & 79\\ 
			\\
			Geometric Similarity Based (Our Method) & $J_{min}=0$ & 4127 & 364\\
			Geometric Similarity Based & $J_{min}=0.05$ & 4127 & 434\\
			Geometric Similarity Based & $J_{min}=0.10$ & 4127 & 490\\
			%Geometric Similarity Based & $J_{min}=0.15$ & 4127 & 549\\
			\bottomrule
		\end{tabular}
		\caption{Destination Regions Extraction Results ($T_{min}=60min, F_{min}=1$)}
		\label{table:DestinationRegionsExtractionResults60min}   
	\end{table*}

	Fig.~\ref{fig:StayRegionExtractionResults} shows a visual perspective of the extracted stay regions by each of the stay region extraction methods in a selected area of the main GPS trajectory. We selected this particular region (Fig.~\ref{fig:MapOfTheSelectedArea_Mapnik}) because it contains clearly depicted places which indicate the car parks. We consider the car park geometries as the ground truth for our empirical observation and geometric similarity analysis. We cropped the GPS trajectory only to cover the selected area by removing all the GPS track point lie outside the geometry of the selected area.
	
	As it is seen in Fig.~\ref{fig:DiameterBasedStays}, the extracted stay regions by diameter based method does not have acceptable results. Although some of the exacted stay regions intersect with the car park regions, they cover the considerable areas outside the car parks. The reference point based stay extraction method~\citep{Zheng:LifePatterns:2009} depicted in Fig.~\ref{fig:ReferencePointBasedStays} has much better performance compared to diameter based method since most of the extracted stay regions intersect with the car parks. However, on the bottom left side of the area, some irrelevant stays are evident. Fig.~\ref{fig:TimeWeightedCentroidBasedStays} shows the extracted stays using our proposed method. Although there are some minor stay regions extracted outside the car parks geometries (in places the same as those in Fig.~\ref{fig:ReferencePointBasedStays}), the extracted stay regions are more compact and more biased to the car parks geometries. 
	
	Table~\ref{table:StayExtractionResults60min} reports that our method has extracted 292 stay regions with only one GPS points (cluster with one member) whereas, the two baseline methods were not able to detect them. The baseline methods compute the value of $\Delta{t}$ as the time distance between two consecutive points because need at least two GPS points in a cluster to make a valid stay region. However, our method incorporates the time-value of the current track point in computing $\Delta{t}$. The time-value of the current track point compensates the cases where the next point lies outside the current stay region ($\Delta{d}>D_{max}$) but the time gap is long enough ($\Delta{t} \ge T_{min}$), resulting in the stay points with only one track point being detected. 
	
	Table~\ref{table:StayExtractionResults60min} also shows that our method has extracted a considerably higher number of stay regions compared to the baselines. The reason is, there might be clusters which have members more than one but the time duration of the stay is less than $T_{min}$ without considering the time-value. In this case, the stay is considered invalid. The duration of the same stay might become more than or equal to $T_{min}$ by considering the time-value of the last point in the cluster. In such scenario, our method detects these clusters while the other two methods miss them. 
	
	In our method, after adding a point to point set of a stay, we update the coordinate of the reference point of the stay by computing the time-weighted centroid of the points in the stay. Therefore, the points extracted in a stay become more biased and closer to the places which longer stops have taken place. Whereas, reference point based method considers the first point of the stay as the reference point does not update it iteratively. The diameter based method~\citep{ProjectLachesis:2004} does not use a centroid point or a reference point and instead uses the diameter of the stay region as the condition of a valid stay. So, it performs much less accurate than both methods.

	\subsection{Destination Extraction Experimental Results}
	\label{DestinationExtractionExperimentalResults}
	
	In this section, we examine and compare our method with two baseline destination extraction methods using the stay regions extracted in the previous section. Table~\ref{table:DestinationRegionsExtractionResults60min} shows the results for three methods. It is evident that the parameters $Diameter_{min}$, $minPts$, and $J_{min}$ have a significant impact on the number of extracted destinations in all three methods. In the diameter based method, the larger $Diameter{min}$ leads to a fewer number of destinations since destinations with the larger area are constructed. The parameter $minPts$ has a significant impact on the number of destinations in density based method. The higher $minPts$ leads to a fewer number of destinations. The greater $J_{min}$ in our method leads to higher number of destinations. 
	
\begin{figure*}[]
		\centering
		\subfigure[t][Destination Grid, Diameter Based ($T_{min}=60min, Diameter_{max}=300m$)]{\label{fig:DiameterBasedDestinations}\includegraphics[width=0.48\linewidth]{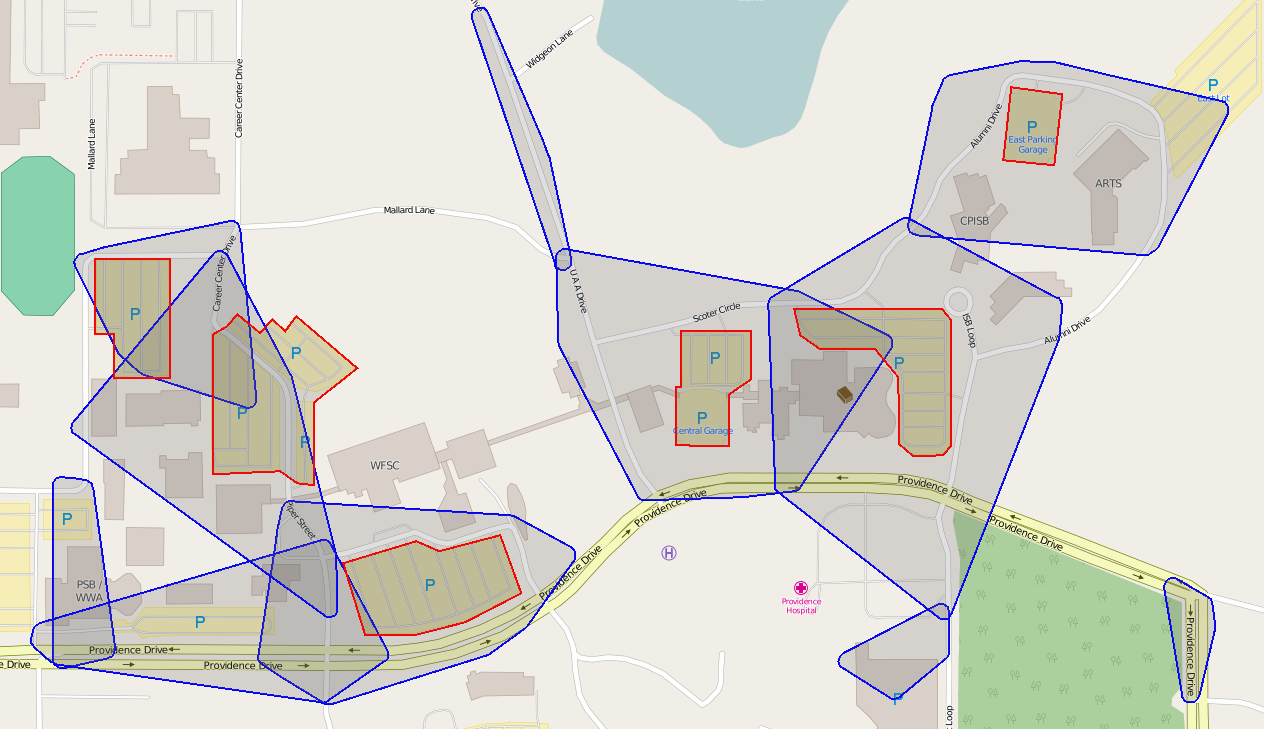}}
		~\hspace{1em}
		\subfigure[t][Destination Grid, Density Based ($T_{min}=60min, D_{max}=100m, minPts=6, eps=100m$)]{\label{fig:DensityBasedDestinations}\includegraphics[width=0.48\linewidth]{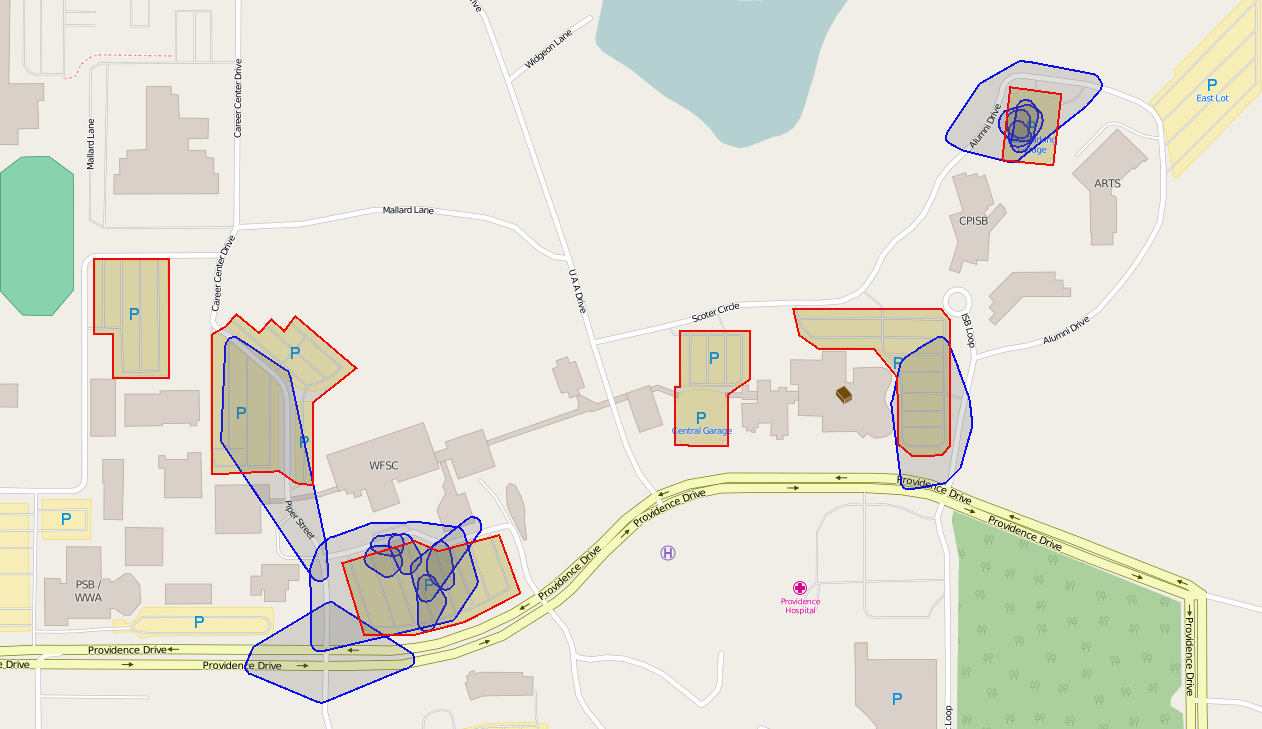}}
		\subfigure[t][Destination Grid, Geometric Similarity Based ($T_{min}=60min, D_{max}=100m, J_{min}=0, F_{min}=1$ )]{\label{fig:GeometricSimilarityBasedDestinationsJ0F1}\includegraphics[width=0.48\linewidth]{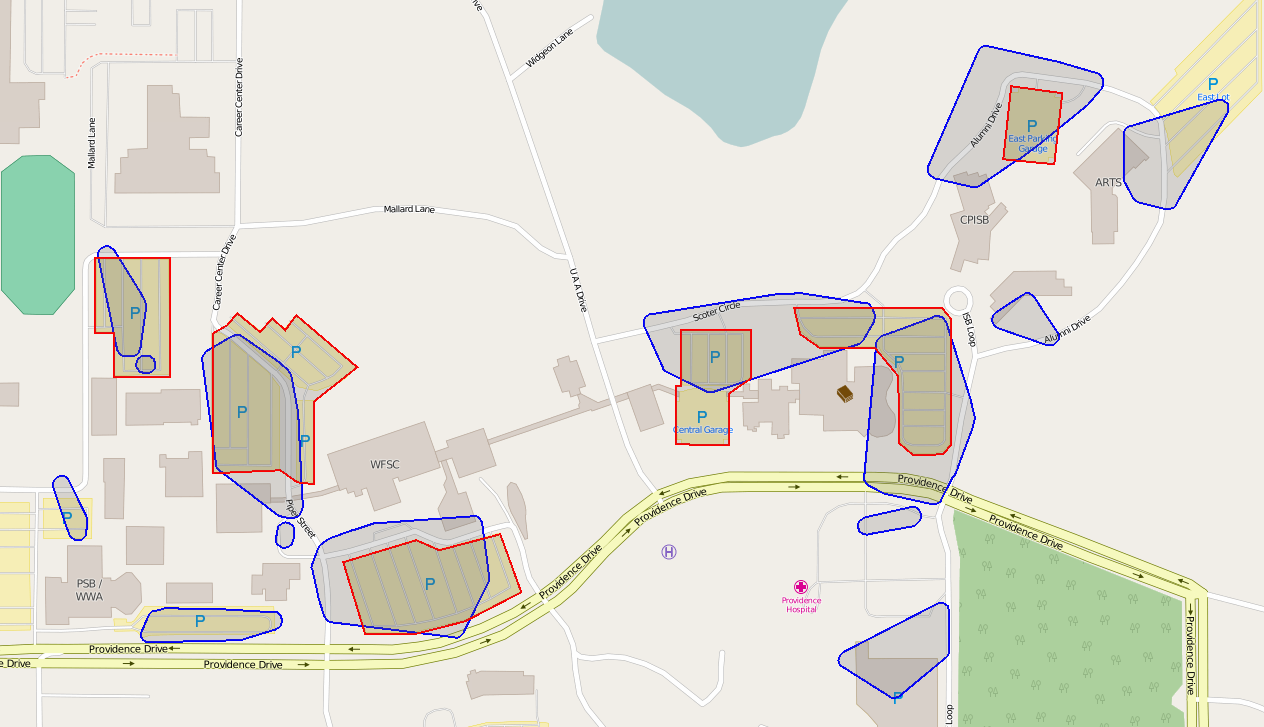}}   
		~\hspace{1em} 
		\subfigure[t][Destination Grid, Geometric Similarity Based ($T_{min}=60min, D_{max}=100m, J_{min}=0, F_{min}=6$)]{\label{fig:GeometricSimilarityBasedDestinationsJ0F6}\includegraphics[width=0.48\linewidth]{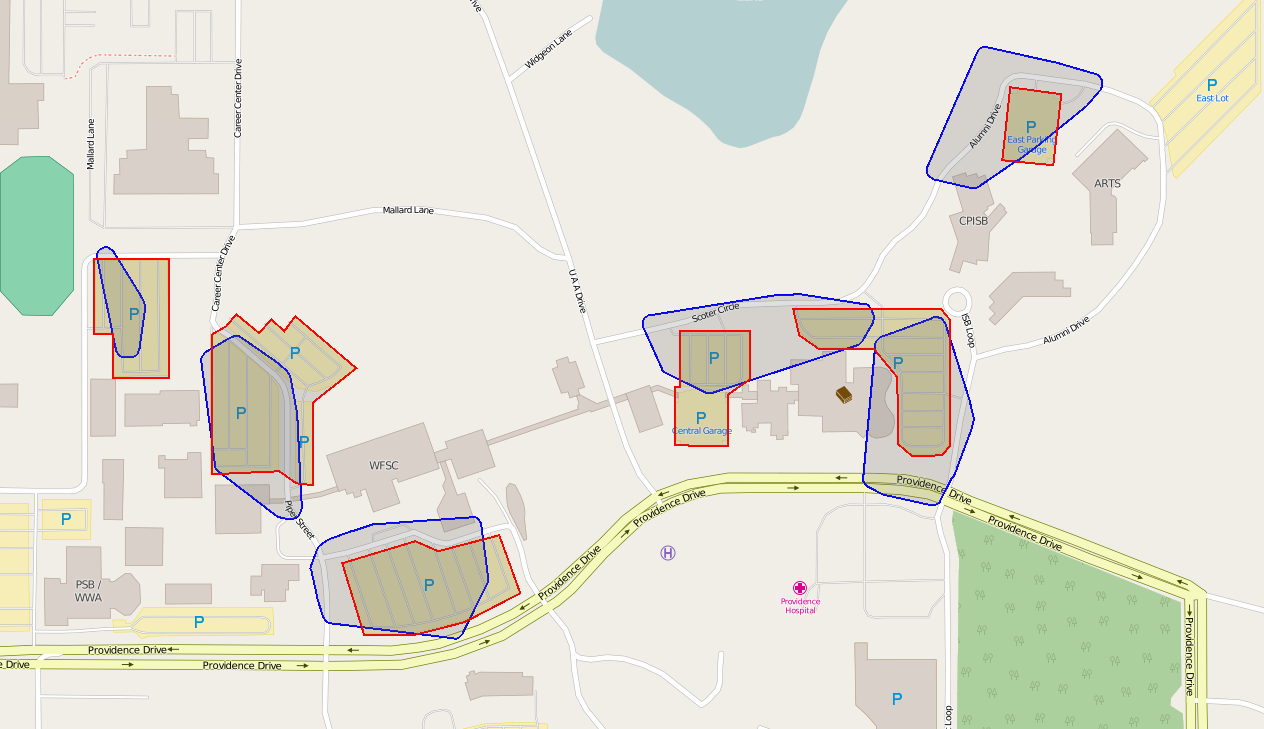}}   
		\subfigure[t][Destination Grid, Geometric Similarity Based ($T_{min}=60min, D_{max}=100m, J_{min}=0.10, F_{min}=1$)]{\label{fig:GeometricSimilarityBasedDestinationsJ01F1}\includegraphics[width=0.48\linewidth]{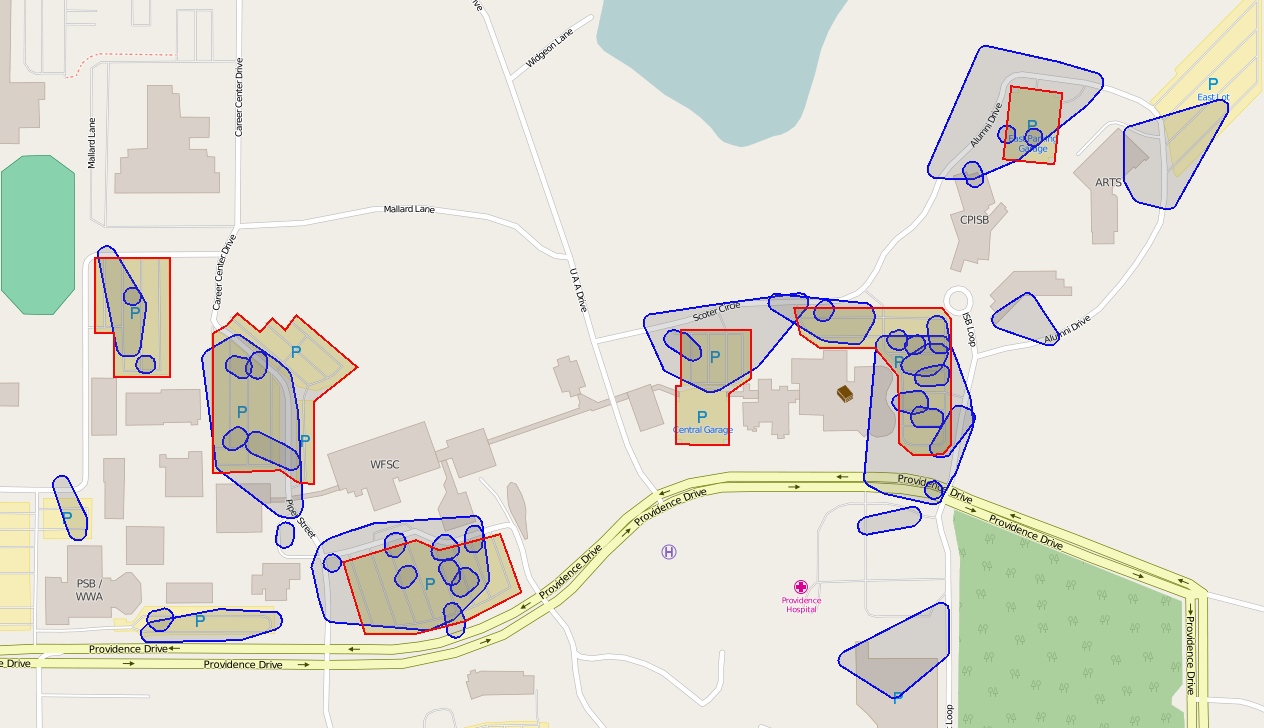}}    
		~\hspace{1em}
		\subfigure[t][Destination Grid, Geometric Similarity Based ($T_{min}=60min, D_{max}=100m, J_{min}=0.10, F_{min}=6$ )]{\label{fig:GeometricSimilarityBasedDestinationsJ01F6}\includegraphics[width=0.48\linewidth]{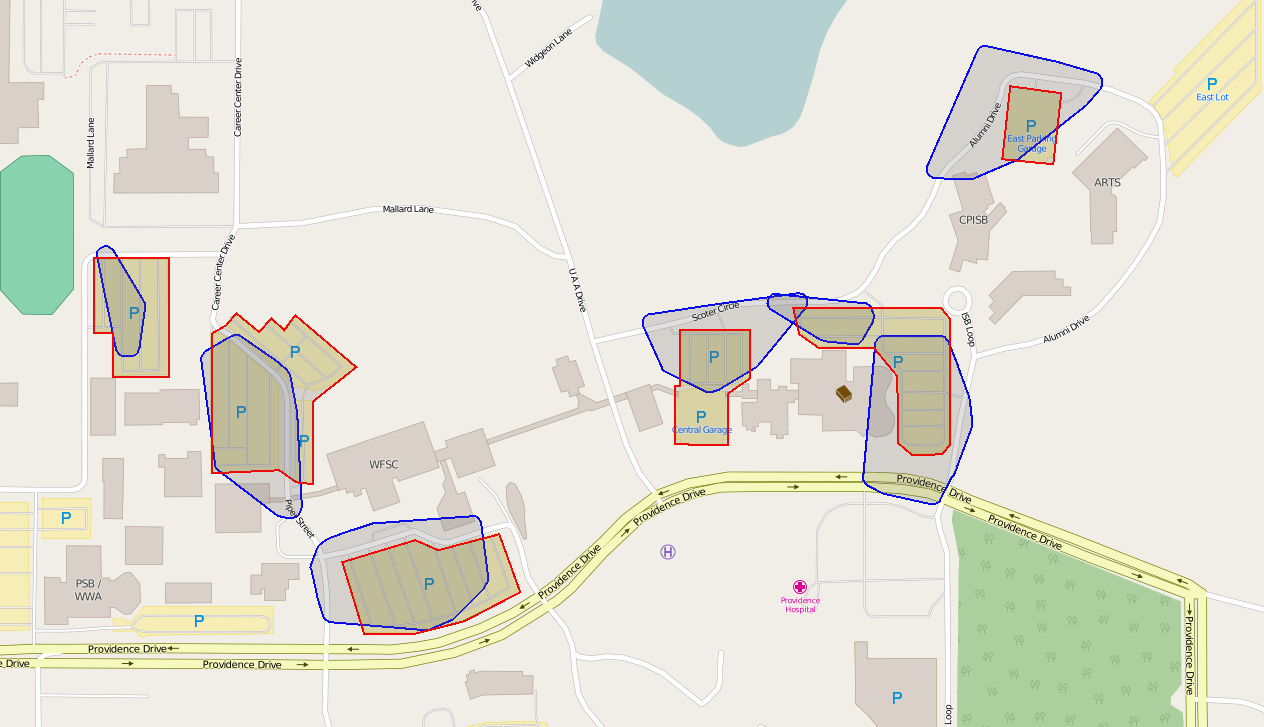}}  
		\subfigure[t][GOI Grid, Geometric Similarity Based ($T_{min}=60min, D_{max}=100m, J_{min}=0.10,  F_{min}=6$)]{\label{fig:TimeWeightedGOIGridJ01F6}\includegraphics[width=0.48\linewidth]{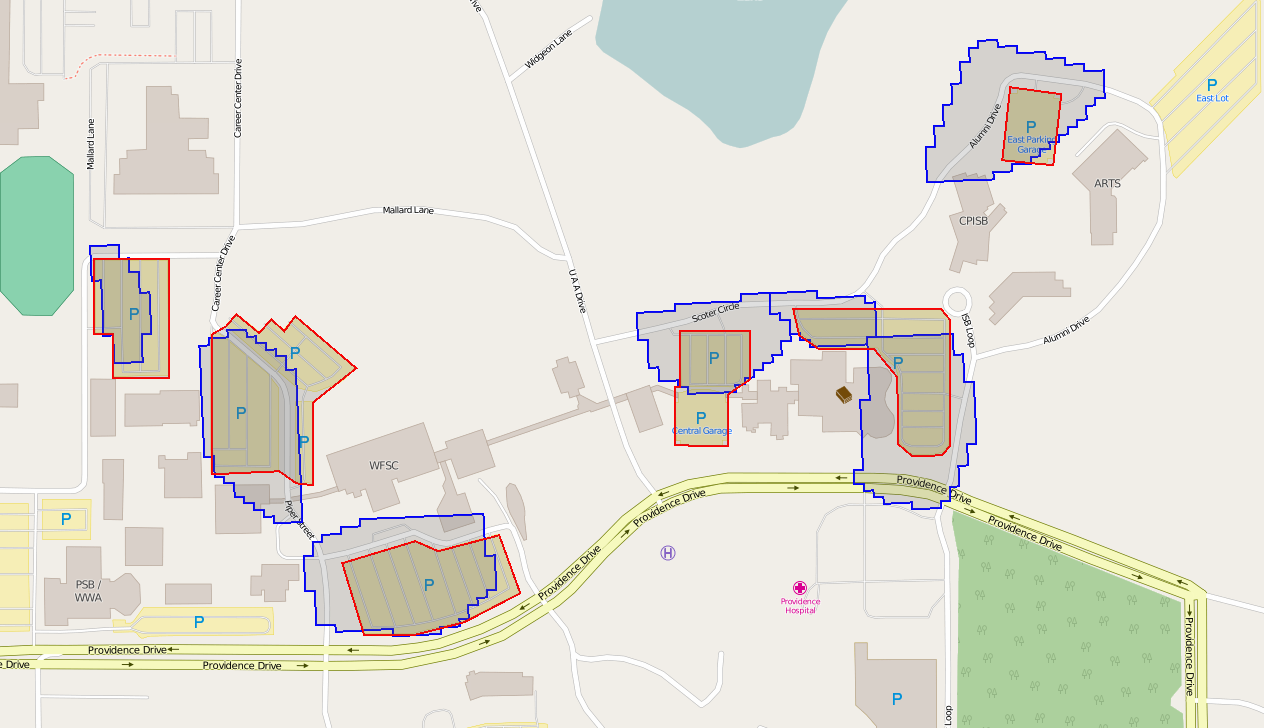}}
		~\hspace{1em}
		\subfigure[t][Final-Grid, Geometric Similarity Based ($T_{min}=60min, D_{max}=100m, J_{min}=0.10, F_{min}=6$)]{\label{fig:TimeWeightedFinalGridJ01F6}\includegraphics[width=0.48\linewidth]{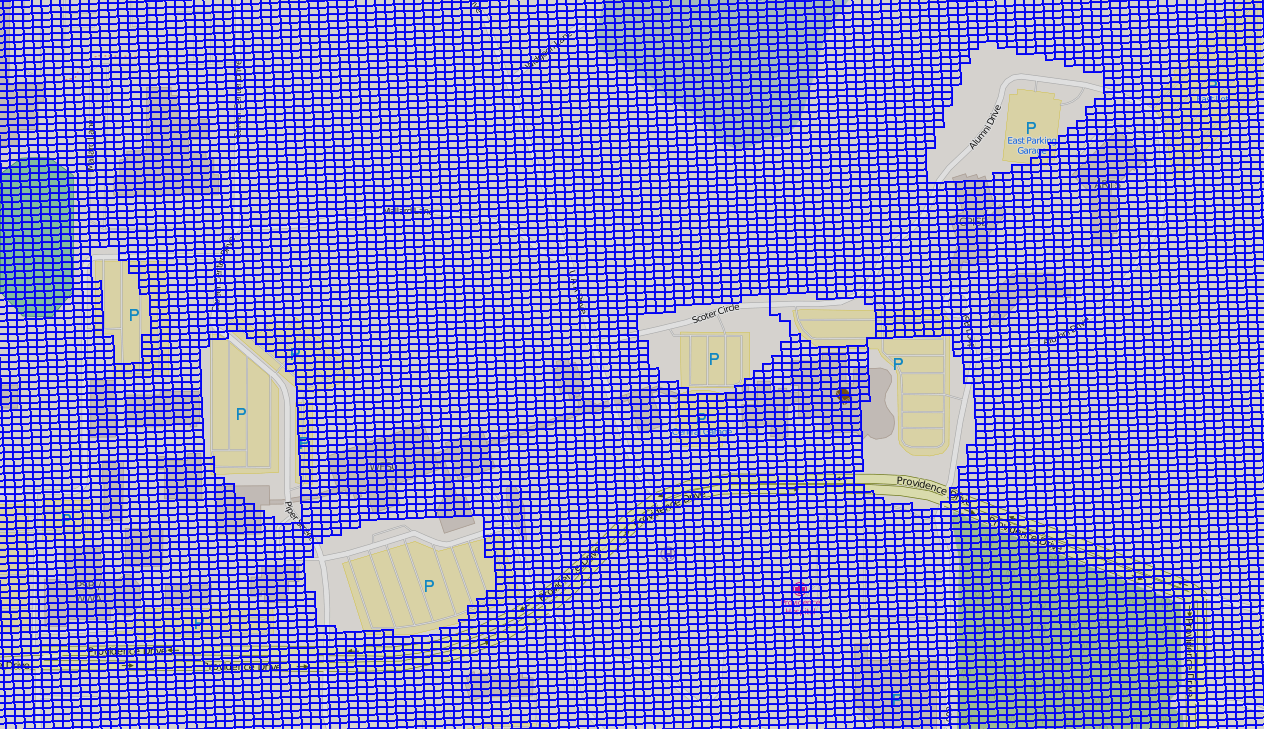}}
		\caption{Destination-Grid, GOI-Grid, and Final-Grid Extraction Results}
		\label{fig:DestinatioGOIandFinalGridExtractionResults}
	\end{figure*}

	Fig.~\ref{fig:DestinatioGOIandFinalGridExtractionResults} shows the results of the destination extraction methods on the map. As it is evident, the diameter based method (Fig.~\ref{fig:DiameterBasedDestinations}) does not have acceptable performance in extracting the geometries of the destinations. Although the extracted destinations do cover the car parks, they have areas much larger than the car park areas, and also, they have significant overlaps. Density based method has more acceptable performance that diameter based method. However, it loses two of the car parks. Moreover, it covers places not related to the car parks. Fig.~\ref{fig:GeometricSimilarityBasedDestinationsJ0F1} shows the destinations extracted by our geometric similarity based method. It is evident our method has constructed destination regions with much more acceptable geometric similarity to the car parks.
	
	\begin{table*}[t!]
		\centering
		\begin{tabular}{ccc} 
			%\toprule
			\cmidrule(r){1-3}
			GOI Extraction Method & Parameters & Degere of Geometric Similarity\\
			\midrule
			Diameter Based & $Diameter_{min}=200m$ & 0.415\\
			Diameter Based & $Diameter_{min}=300m$ & 0.097\\
			Diameter Based & $Diameter_{min}=400m$ & 0.130\\
			%Diameter Based & $Diameter_{min}=500m$ & 0.053\\
			%Diameter Based & $Diameter_{min}=600m$ & 0.050\\
			\\    
			Density Based & $eps=100m,minPts=3$ & 0.310\\
			Density Based & $eps=100m,minPts=6$ & 0.329\\
			Density Based & $eps=100m,minPts=9$ & 0.314\\
			%Density Based & $eps=100m,minPts=12$ & 0.014\\
			%Density Based & $eps=100m,minPts=15$ & 0.028\\
			\\
			Geometric Similarity Based & $J_{min}=0$ & 0.623\\
			Geometric Similarity Based & $J_{min}=0.05$ & 0.628\\
			Geometric Similarity Based & $J_{min}=0.10$ & 0.650\\
			%Geometric Similarity Based & $J_{min}=0.15$ & 0.668\\
			%Geometric Similarity Based & $J_{min}=0.20$ & 0.375\\
			\bottomrule
		\end{tabular}
		\caption{GOI Extraction Extraction Results ($T_{min}=60min, F_{min}=6$)}
		\label{table:GOIExtractionResults60min}   
	\end{table*}   
	
	As it is evident in figures~\ref{fig:GeometricSimilarityBasedDestinationsJ0F1} and~\ref{fig:GeometricSimilarityBasedDestinationsJ01F1}, in our method with with $F_{min}=1$, the value $J_{min}=0$ leads to all the geometries of all the destinations being disjoint while the number of extracted destination using $J_{min}=0.1$ is much higher. The destinations are overlapping, and even some destinations are fully covered by the other destinations.
	
	Figures~\ref{fig:GeometricSimilarityBasedDestinationsJ0F6} and~\ref{fig:GeometricSimilarityBasedDestinationsJ01F6} illustrate the extracted destinations with $F_{min}=6$. Comparison of the figures with figures~\ref{fig:GeometricSimilarityBasedDestinationsJ0F1} and~\ref{fig:GeometricSimilarityBasedDestinationsJ01F1} clearly indicates the effect of parameter $F_{min}$ in our destination extraction method. In the latter figures, the destinations with fewer visit frequencies have been eliminated from the destination-grids, and only the destinations which have more geometric similarity to car park areas have been left.
	
	Comparing two figures~\ref{fig:GeometricSimilarityBasedDestinationsJ0F6} and~\ref{fig:GeometricSimilarityBasedDestinationsJ01F6} reveals the effect of the value of $J_{min}$ on the extracted destinations. The extracted destinations in both figures are quite similar except for the destination in the middle of the area. Fig.~\ref{fig:GeometricSimilarityBasedDestinationsJ0F6} has merged the area of the two neighboring car parks together while Fig.~\ref{fig:GeometricSimilarityBasedDestinationsJ01F6} has extracted two distinct geometries for the same destination. This shows the better performance of the method with parameter $J_{min}=0.1$.
	
	\subsection{Paritioning Experimental Results}
	\label{PartitioningExperimentalResults}
	
	The GOI-Grid, which is the result of constructing the GOIs based on the destinations in Fig.~\ref{fig:GeometricSimilarityBasedDestinationsJ01F6} is illustrated in Fig.~\ref{fig:TimeWeightedGOIGridJ01F6}. It is clearly seen that the partitioning method has resolved the problem of two destinations having a geometric overlap. The two destination regions in the middle of the Fig.~\ref{fig:GeometricSimilarityBasedDestinationsJ01F6} have been partitioned into two distinct cells in GOI-Grid without having any intersection. 
	
	The Final-grid, which is depicted in Fig.~\ref{fig:TimeWeightedFinalGridJ01F6} is the last result of our partitioning. It is evident that the final-grid guarantees both characteristics of a valid partition. None of the cells overlap each other and all the GPS points in the mobile object trajectory can be labeled with the ID of a cell in the final grid. 
	
	\subsection{Geometric Similarity Evaluation Results}
	\label{GeometricSimilarityEvaluationResults}
	
	In this section, we use the geometric similarity as a quantitive metric to analyze the quality of our partitioning method compared to the baselines. We use Eq.~\ref{eq:GeometricSimilarity} for analyzing the performance of the similarity of the real GOIs and the estimated GOIs. This metric uses the proportion of the area of the intersection of two geometries ($g_i$ and $g_j$) to the area of the union of them. 
	
	Table~\ref{table:GOIExtractionResults60min}, presents the calculated degree of geometric similarity between the geometries of the real GOIs (the red colored polygons in figure~\ref{fig:TimeWeightedGOIGridJ01F6}) and their corresponding extracted GOIs (the blue colored polygons in figure~\ref{fig:TimeWeightedGOIGridJ01F6}). As it is evident, our method has the highest values for the geometric similarity. Table~\ref{table:GOIExtractionResults60min} also shows that the values of geometric similarities vary based on different values for the parameters $J_{min}$, $minPts$, and $Diamater_{min}$. 
	
	Among the evaluated methods with different parameters, our method with $J_{min}=0.15$ has the best results. Therefore, the partitioning method which maximizes the geometric similarity (discussed in section~\ref{ProblemDefinition}) is our method with $J_{min}=0.15$.  
	
	\section{Conclusion and Future Work}
	\label{ConclusionAndFutureWork}
	In this paper, we addressed the problem of finding the Geometries of Interest of a mobile object and partitioning the trajectory area into a grid through analyzing its GPS trajectories. The research shows that considering the concept of time-value of the GPS points significantly improves the accuracy of stay region extraction. Moreover, the results of this study support the idea that considering the geometries of the stay regions, makes the geometries estimated GOIs remarkably more similar to the real world GOIs. 
	
	This research has opened up many questions in need of further investigation and will serve as a base for future studies. It would be interesting to focus on improving the performance and the accuracy of our proposed partitioning method by aggregating the trajectory data of other mobile objects moving in the same area of our particular mobile object. Other improvements such as using outlier detection methods to detect and remove the outlier points from the point set of the destinations could improve the results further. Finding the best geometric similarity metric to improve the performance of the destination extraction phase would be another interesting research problem that could be addressed in the future.
	
	\bibliographystyle{spbasic} 
	\bibliography{thesis}
	
\end{sloppypar}

\end{document}